\documentclass{article}

\usepackage[utf8]{inputenc} % allow utf-8 input
\usepackage[T1]{fontenc}    % use 8-bit T1 fonts
\usepackage{hyperref}       % hyperlinks
\usepackage{url}            % simple URL typesetting
\usepackage{booktabs}       % professional-quality tables
\usepackage{amsfonts}       % blackboard math symbols
\usepackage{nicefrac}       % compact symbols for 1/2, etc.
\usepackage{microtype}      % microtypography
\usepackage{xcolor}         % colors

\usepackage{amsmath}
\usepackage{amssymb}
\usepackage{amsthm}

\usepackage{physics}
\usepackage{algorithm}
\usepackage{algorithmicx}
\usepackage{algpseudocode}

\usepackage{graphicx}
\usepackage{caption}
\usepackage{subcaption}
\usepackage{numprint}
\usepackage{appendix}

\usepackage{authblk}

\usepackage{geometry}
\usepackage[
    backend=biber,
    style=authoryear,
    sorting=nyt,
    citestyle=authoryear,
    natbib=true,
    doi=false,isbn=false,url=false,eprint=false,
    maxcitenames=2,
    maxbibnames=10,
    dashed=false
]{biblatex}
\defbibfilter{appendixOnlyFilter}{
  segment=1 % Segment 1 will be chosen to be the one in appendix
  and not segment=0 % Default segment is 0
}

\theoremstyle{definition}

\newtheorem{definition}{Definition}[section]
\newtheorem{theorem}{Theorem}[section]
\newtheorem{proposition}{Proposition}[section]

\newcommand{\Exp}[2]{\mathbb{E}_{#1}\left[ #2 \right]}
\DeclareMathOperator{\VarOp}{Var}
\newcommand{\Var}[2]{\VarOp_{#1}\left[ #2 \right]}

\newcommand{\bx}{\boldsymbol x}

\newcommand{\bX}{\boldsymbol X}

\newcommand{\bg}{\boldsymbol g}

\newcommand{\by}{\boldsymbol y}

\newcommand{\btheta}{\boldsymbol \theta}

\newcommand{\bpi}{\boldsymbol \pi}
\newcommand{\bw}{\boldsymbol w}

\newcommand{\bxi}{\boldsymbol\xi}

\newcommand{\data}{D} % data
\newcommand{\neigh}{D'} % neighbouring data

\newcommand{\params}{\btheta} % model parameters

\newcommand{\elbo}{\mathcal{L}}

\newcommand{\qparams}{\bxi}
\newcommand{\qm}{\boldsymbol{m}_q} % mean
\newcommand{\qs}{\boldsymbol{s}_q} % scale
\newcommand{\qsig}{\boldsymbol{\sigma}_q} % variance
\newcommand{\draw}{\boldsymbol{\eta}} % sample from std normal used in reparametrization trick
\newcommand{\qa}{\boldsymbol{a}_q} % scale

\newcommand{\dpdraw}{\boldsymbol{\psi}}

\newcommand{\gm}{\nabla_{\qm}}
\newcommand{\gs}{\nabla_{\qs}}
\newcommand{\gparams}{\nabla_{\params}}
\newcommand{\gqm}{\boldsymbol{g}_m}
\newcommand{\gqs}{\boldsymbol{g}_s}
\newcommand{\gq}{\boldsymbol{g}}
\newcommand{\gqmx}{\boldsymbol{g}_{m,\bx}}
\newcommand{\gqsx}{\boldsymbol{g}_{s,\bx}}
\newcommand{\gqx}{\boldsymbol{g}_{\bx}}
\newcommand{\dpgqm}{\tilde{\boldsymbol{g}}_m}
\newcommand{\dpgqs}{\tilde{\boldsymbol{g}}_s}
\newcommand{\dpgq}{\tilde{\gq}}
\newcommand{\gqa}{\boldsymbol{g}_a}

\newcommand{\Tbo}{T_{\text{burn-out}}}
\graphicspath{{./figures/}}

\title{DPVIm: Differentially Private Variational Inference Improved}

\author[1]{Joonas Jälkö}
\author[1]{Lukas Prediger}
\author[2]{Antti Honkela}
\author[1,3]{Samuel Kaski}
\affil[1]{\scriptsize Helsinki Institute for Information Technology HIIT, Department of Computer Science, Aalto University}
\affil[2]{\scriptsize Helsinki Institute for Information Technology HIIT, Department of Computer Science, University of Helsinki}
\affil[3]{\scriptsize Department of Computer Science, University of Manchester}

\date{}

\begin{document}

\maketitle
%%%%%%%%%%%%%%%%%%%%%%%%%%%%%%%%%%%%%%%%%%%%%%%%%%%%%%%%%%%%%%

\begin{abstract}
    Differentially private (DP) release of multidimensional statistics typically considers an aggregate sensitivity, e.g. the vector norm of a high-dimensional vector. However, different dimensions of that vector might have widely different magnitudes and therefore DP perturbation disproportionately affects the signal across dimensions.
    We observe this problem in the gradient release of the DP-SGD algorithm when using it for variational inference (VI), where it manifests in poor convergence as well as high variance in outputs for certain variational parameters, and make the following contributions:
    (i) We mathematically isolate the cause for the difference in magnitudes between gradient parts corresponding to different variational parameters. Using this as prior knowledge we establish a link between the gradients of the variational parameters, and propose an efficient while simple fix for the problem to obtain a less noisy gradient estimator, which we call \emph{aligned} gradients. 
    This approach allows us to obtain the updates for the covariance parameter of a Gaussian posterior approximation without a privacy cost. 
    We compare this to alternative approaches for scaling the gradients using analytically derived preconditioning, e.g. natural gradients.
    (ii) We suggest using iterate averaging over the DP parameter traces recovered during the training, to reduce the DP-induced noise in parameter estimates at no additional cost in privacy.
    Finally, (iii) to accurately capture the additional uncertainty DP introduces to the model parameters, we infer the DP-induced noise from the parameter traces and include that in the learned posteriors to make them \emph{noise aware}.
    We demonstrate the efficacy of our proposed improvements through various experiments on real data.
\end{abstract}

%%%%%%%%%%%%%%%%%%%%%%%%%%%%%%%%%%%%%%%%%%%%%%%%%%%%%%%%%%%%%%

\section{Introduction}

Differential privacy (DP) \citep{DworkMNS06} protects privacy of data subjects by limiting how much about the input data can be learned from the output of an algorithm. Additive noise mechanisms achieve DP by adding noise calibrated to the maximum change in function output due to a single individual, known as sensitivity. When releasing high-dimensional data through such mechanisms, different variables may have widely different sensitivities, which is typically ignored or unknown. Instead, the sensitivity of the release is computed as an aggregate over all the dimensions, which we call \emph{global sensitivity}, in contrast to variable-specific \emph{local sensitivity}. As the DP noise is scaled with this global sensitivity, it affects dimensions with lower local sensitivities more. A prominent example where this occurs is the gradient release in DP stochastic gradient descent (DP-SGD) \citep{SongCS13, BassilyST14, Abadi+16}. Furthermore, the released final parameters from DP-SGD are noisy estimators of the true parameters and the noise is treated as an inherent trade-off of providing privacy \citep{Abadi+16}. In combination, the large amounts of non-quantified noise can lead to large errors when the noisy results are deployed, as well as make them extremely non-robust.

We discover these issues in the perturbed gradients used in DP variational inference (DPVI) \citep{JalkoDH17} where they lead to severe errors in capturing posterior uncertainty. This results e.g. in poor predictive uncertainty estimation, making the predictions of the learned model less robust. DPVI is a widely applicable state-of-the-art algorithm for privacy-preserving (approximate) Bayesian inference, based on DP-SGD.

We mathematically isolate the cause for these problems in DPVI, and propose and evaluate two ways of alleviating the problem of gradient scales in DPVI: one scales gradients with a preconditioning matrix before applying the DP mechanism, the other is based on insights into the mathematical structure of the gradients, which reveals that their components are mathematically linked and can be derived from each other in a post-processing step.

Additionally, we theoretically and experimentally evaluate the method of iterate averaging as a way to further improve the parameter estimate as well as approximate the additional variance induced by DP perturbations to DPVI to make the posterior approximation noise aware at no additional cost in privacy.

%%%%%%%%%%%%%%%%%%%%%%%%%%%%%%%%%%%%%%%%%%%%%%%%%%%%%%%%%%%%%%

\subsection{Related work}

In the context of DP-SGD, the following previous works acknowledge the different local sensitivities of different parts of the full gradient: \cite{McMahanRTZ17} suggested clipping the gradients of a neural network separately for each layer to avoid the clipping-induced bias \citep{ChenWH20}. This approach was further improved by \cite{AndrewTMR21} by learning the per-layer clipping threshold from the gradients under DP. However, since the perturbation is still scaled with the total sensitivity aggregated over the dimensions, this approach does not improve the disparate effect that the Gaussian noise will have on the dimensions with smaller gradients.

For noise aware DP Bayesian inference, the most related work is by Bernstein and Sheldon \citep{BernsteinS18, BernsteinS19} and \cite{KulkarniJKKH21}. These works include the DP perturbation mechanism into a probabilistic model using perturbed sufficient statistics as the inputs. This allows capturing the DP-induced additional uncertainty in the posterior distribution of model parameters.

%%%%%%%%%%%%%%%%%%%%%%%%%%%%%%%%%%%%%%%%%%%%%%%%%%%%%%%%%%%%%%

\section{Preliminaries}

\subsection{Differential privacy}

\begin{definition}
For $\epsilon \geq 0$ and $\delta \in [0,1]$, a randomised mechanism $\mathcal{M}$ satisfies $(\epsilon,\delta)$-differential privacy~\citep{DworkMNS06} if for any two data sets different in only one element, $\data, \neigh \in \mathcal{D}$, and for all outputs $S \subseteq \text{im}(\mathcal{M})$, the following constraint holds: 
\begin{equation}\label{eq:dp}
    \Pr(\mathcal{M}(\data) \in S) \leq  \exp(\epsilon) \Pr(\mathcal{M}(\neigh) \in S)  + \delta.
\end{equation}
\end{definition}

%%%%%%%%%%%%%%%%%%%%%%%%%%%%%%%%%%%%%%%%%%%%%%%%%%%%%%%%%%%%%%

\subsection{Variational inference}
\label{sec:variational_bayes}

Variational inference is a commonly applied technique in probabilistic inference, where the aim is to learn an approximation for a (typically intractable) posterior distribution of the parameters of a probabilistic model \citep{JordanGJS99}. This is done by maximising a quantity called \emph{evidence lower bound} (ELBO) over the parameters of the variational approximation. For a probabilistic model $p(\data, \params)$, where $\data$ denotes the observed variables and $\params$ the model parameters, and for a variational approximation $q(\params)$ of the posterior, the ELBO is given as
\begin{align}
    \elbo(q) 
    &= \log p(D) - \text{KL}(q(\params) \, || \, p(\params \mid \data)) \\
    &= \Exp{q(\params)}{\log p(\data, \params)} + H(q),
    \label{elbo}
\end{align}
where $\text{KL}$ denotes the Kullback-Leibler divergence and $H$ the (differential) entropy.

In the following we first restrict ourselves to the commonly used \emph{mean-field} variational inference, i.e., using a Gaussian with diagonal covariance as the posterior approximation. We will later generalise this to a full-rank coveriance approximation. For $d$-dimensional data the diagonal approximation is parametrised by the means $\qm \in \mathbb{R}^d$ and the dimension-wise standard deviations $\qsig \in \mathbb{R}^d$. We further reparametrise the model with $\qs = T^{-1}(\qsig)$, where $T: \mathbb{R} \rightarrow \mathbb{R}_+$ is monotonic, in order to facilitate optimisation in an unconstrained domain. Both $T$ and $T^{-1}$ are applied element-wise for each of the parameters. Common choices for $T$ are the exponential function $T(s) = \exp(s)$ and the softplus function $T(s) = \log(\exp(s) + 1)$ (used e.g. in the Pyro probabilistic programming package \citep{Bingham+19}). We use $\qparams = (\qm, \qs)$ to refer to the complete set of variational parameters.

A draw from this posterior distribution can then be written as \citep{KingmaW13}:
\begin{align}
    \params := \params(\draw; \qm, \qs) = \qm + T(\qs)\draw,
    \label{eq:reparam}
\end{align}
where $\draw \sim N(0, I_d)$, $\params, \draw \in \mathbb{R}^d$ and $I_d$ is a $d$-dimensional identity matrix. 
\cite{KucukelbirTRGB17} use this reparametrisation trick together with single-sample MC integration to give the ELBO a differentiable form with gradients:
\begin{align}
    \gqm :&= \gm\elbo(q) \nonumber \\
          &= \gm \log p(\data, \params(\draw; \qm, \qs))  \label{eq:mc_m_grad}\\
    \gqs :&= \gs\elbo(q) \nonumber \\
          &= \gs \log p(\data, \params(\draw; \qm, \qs)) + \gs H(q) \label{eq:mc_s_grad},
\end{align}
where $\draw \sim N(0, I)$. Throughout this work we assume that the likelihood factorises as: $p(\data \mid \params) = \prod_{\bx \in \data} p(\bx \mid \params)$. Using $N$ to denote the size of $\data$, we can now further decompose the gradients in \eqref{eq:mc_m_grad} and
\eqref{eq:mc_s_grad} as 
\begin{align}
    \gqm &= \sum_{\bx \in \data} \left(\gm \log p(\bx \mid \params(\draw; \qm, \qs))  \right. \nonumber \\
                              &\left.  + \frac{1}{N} \gm \log p(\params(\draw; \qm, \qs))\right) \label{mc_m_grad_sum}\\
    %\gqm &= \sum_{\bx \in \data} \gqmx \label{mc_m_grad_sum} \\
    \gqs &= \sum_{\bx \in \data} \left(\gs \log p(\bx \mid \params(\draw; \qm, \qs))  \right.  \nonumber \\
                              &\left.  +\frac{1}{N}\left(\gs \log p(\params(\draw; \qm, \qs)) 
                                        + \gs H(q) \label{mc_s_grad_sum}\right)\right).
\end{align}
We denote the per-example gradient components (i.e., those for each individual $\bx$) that appear in the above sums with $\gqmx$ and $\gqsx$ respectively.

A common approach to performing variational inference in practice is to initialise $\qs$ to small values, which allows the algorithm to move $\qm$ quickly close to their optimal values due to large error in the KL term of the ELBO induced by the narrow approximation.

%%%%%%%%%%%%%%%%%%%%%%%%%%%%%%%%%%%%%%%%%%%%%%%%%%%%%%%%%%%%

\section{Differentially private variational inference}

The first algorithm for differentially private variational inference for non-conjugate models \citep{JalkoDH17} optimises the ELBO using gradients \eqref{eq:mc_m_grad} and \eqref{eq:mc_s_grad} with differentially private stochastic gradient descent \citep{Abadi+16} to provide privacy. This involves concatenating each of the per-example gradients to obtain $\gqx = (\gqmx^T, \gqsx^T)^T$, clipping $\gqx$ so that it has $\ell_2$ norm no larger than a threshold $C$ to limit the sensitivity, and finally adding Gaussian noise to the sum of these clipped per-example gradients to obtain $\dpgq$, which is used for the parameter update. We refer to this algorithm in the following as \emph{vanilla DPVI}.

This formulation induces a problem which, while seemingly minor at first glance, severely affects accuracy of solutions. We next isolate this problem and then propose a solution through detailed analysis on the gradients of the variational parameters.

%%%%%%%%%%%%%%%%%%%%%%%%%%%%%%%%%%%%%%%%%%%%%%%%%%%%%%%%%%%%

\subsection{Disparate perturbation of variational parameter gradients}
\label{sec:disparate_perturbation}

While the clipping of the gradients allows us to bound the global sensitivity of the gradient vector, it completely ignores any differences in gradient magnitudes across the dimensions. As DPVI (and more generally DP-SGD) proceeds to add Gaussian noise with standard deviation proportional to the clipping threshold to \emph{all of the dimensions}, the signal-to-noise-ratio can vary greatly across the parameter dimensions. Parameter dimensions that experience low signal-to-noise ratio will converge much slower than others (cf.~\citealp{Domke19} and references therein).\footnote{We also provide a high-level argument why this is the case in Appendix~\ref{app:larger_noise_slows_convergence}.} Next, we will show that such a magnitude difference arises between the gradients of variational parameters $\qm$ and $\qs$.

Note that the gradient of Equation \eqref{eq:reparam} w.r.t.~$\qm$ is $\gm \params(\draw; \qm, \qs) = 1$, which leads to the following proposition (a more detailed derivation can be found in Appendix~\ref{app:proposition_argument}):
\begin{proposition}
\label{thm:gs_from_gm}
    Assume $q$ to be diagonal Gaussian, then the gradient $\gqs$ in Equation \eqref{eq:mc_s_grad} becomes
    \begin{align}
        \gqs &= \draw T'(\qs) \gqm  + \gs H(q), \label{eq:mc_s_grad_as_theta}
    \end{align}
    where $T'$ denotes the derivative of $T$.
\end{proposition}
As the entropy term is independent of the data, our update equation for $\qs$ depends on the data only through $\gqm$. Thus, in order to show that this term gets affected by the noise more than $\gqm$ itself, it suffices to inspect the magnitudes of $\draw$ and $T'(\qs)$. As $\draw \sim N(0, I_d)$, we have $\draw_j = O(1)$. We have $T'(\qs) \leq \qsig $ for common choices of $T$ discussed above (a proof for softplus can be found in Appendix \ref{app:T_proof}).
For the data dependent part of $\gqs$ it then follows that
\begin{align}
    |\draw T'(\qs) \gqm| = O(\qsig) |\gqm|.
\end{align}
Therefore, it is easy to see that as $\qsig$ becomes small, the gradient $\gqs$ becomes small compared to $\gqm$. Note that this is especially problematic combined with the practice of initialising $\qs$ to small values to speed up the convergence of $\qm$, discussed in Sec.~\ref{sec:variational_bayes}.

%%%%%%%%%%%%%%%%%%%%%%%%%%%%%%%%%%%%%%%%%%%%%%%%%%%%%%%%%%%%

\subsubsection{Addressing the differing gradient scales}
\label{sec:dpvi_variants}
Above we have identified a magnitude difference between the gradient components, which leads to variational standard deviation parameters being disproportionately affected by DP noise. Next we use this structural knowledge to propose a method for scaling the gradients to more closely matching magnitudes, after discussing two alternatives based on standard techniques.

\paragraph{Natural Gradients}
As a first solution, we consider \emph{natural gradients} \citep{AmariS98}. This is a common approach for improving convergence for VI (cf.~e.g.~\citealp{Honkela+10, KhanMN18, SalimbeniHEH18}), which relies on scaling the gradients using the information geometry of the optimisation problem.

The natural gradients $\bg^{nat}$ are computed using the inverse of the Fisher information matrix $\mathcal{I}$ as
\begin{align}
    \mathcal{I} &= \Exp{\params|\qs,\qm}{(\gparams \log q(\params)) (\gparams \log q(\params))^T} \\
    \bg^{nat} &= \mathcal{I}^{-1} \bg. \label{eq:nat_grad}
\end{align}
For our setting this leads to
\begin{align}
    \gqm^{nat} &= T(\qs)^2 \gm \elbo(q) \label{eq:g_m_nat} \\
    \gqs^{nat} &= \frac{1}{2T'(\qs)} \left( \draw \gqm^{nat} + \frac{T(\qs)^2}{T'(\qs)} \gs H(q) \right). \label{eq:g_s_nat}
\end{align}

We observe that in the natural gradients the scaling by $T'(\qs)$ in the gradients of $\qs$ is now reversed, meaning that for small $T'(\qs)$ the gradients of $\qs$ will tend to dominate over those of $\qm$. Therefore we expect natural gradients to result in a different instance of the problem of disproportionate DP noise instead of resolving it.

\paragraph{Preconditioning of Gradients}
The simplest way to fix the disproportionate DP noise is preconditioning of the gradients to undo the downscaling of the data-dependent part in Eq.~\eqref{eq:mc_s_grad_as_theta}, by multiplying with $\left(T'(\qs)\right)^{-1}$, to obtain
\begin{equation}
    \gqs^{precon} = \frac{1}{T'(\qs)} \gqs = \draw \gqm + \frac{\gs H(q)}{T'(\qs)}.
\end{equation}
We can see that the data dependent part of $\gqs^{precon}$ (the first term) is of the same magnitude as $\gqm$, and thus the noise affects the gradient components equally.\footnote{Note that this scaling also affects the data-independent entropy term in the gradient for $\qs$. While the scaling term $\left(T'(\qs)\right)^{-1}$ does increase the entropy part for small $\qs$, the data-dependent term is still typically much larger and will dominate the gradient.}

\paragraph{Aligned Gradients}
The preconditioning approach fixed the issue of different magnitudes in the gradients. However,
it comes at the cost of increased $\ell_2$-norm of the full gradient, which requires a higher clipping threshold and therefore increases DP noise. We will now discuss a new alternative method for resolving the disproportionate DP noise problem that avoids this issue.

Equation \eqref{eq:mc_s_grad_as_theta} shows that we can write $\gqs$ in terms of $\gqm$ and an additional entropy term. Since neither the scaling factor $\draw T'(\qs)$ nor the entropy gradient $\gs H(q)$ depend on the data $\data$, it suffices to  release the gradients $\gqm$ under DP as $\dpgqm$, from which we obtain the $\dpgqs$ via Eq.~$\eqref{eq:mc_s_grad_as_theta}$. As this is simply post-processing, it does not incur additional DP cost. 
Because $\dpgqs$ is now computed directly as a transformation of $\dpgqm$, the noise term in both gradients is aligned in proportion to the gradient signals. We refer to this approach as \emph{aligned DPVI} for the rest of the paper. The procedure for computing the aligned DPVI gradients is summarized in Algorithm \ref{alg:aligned}.

\begin{algorithm}[h]
    \begin{algorithmic}
    \State $\params \gets \qm + \draw T(\qs)$ where $\draw \sim \mathcal{N}(0,I)$
    \State $\gqmx \gets \gm \elbo(q)$ for $\bx \in \data$ %\Comment{Compute the per-example gradients for $\qm$}
    \State $\gamma_{\bx} \gets \min(1, \nicefrac{C}{||\gqmx||})$ for $\bx \in \data$ \Comment{clipping factor} %\Comment{Compute the clipping multiplier}
    \State $\dpgqm \gets \sum_{\bx \in \data} \gamma_{\bx} \gqmx + \sigma_{DP} C \dpdraw$, where $\dpdraw \sim \mathcal{N}(0,I)$ %\Comment{Get DP release for $\gqm$}
    \State $\dpgqs^{aligned} \gets \draw T'(\qs) \dpgqm + \gs H(q)$ %\Comment{Get DP aligned $\gqs$ via post-processing}
    \end{algorithmic}
    \caption{The aligned gradient procedure \label{alg:aligned}}
\end{algorithm}

The following theorem (proved in Appendix~\ref{app:proof_aligned_gradients_smaller_scale}) guarantees that the variance in the gradients of $\qs$ is reduced in aligned DPVI:
\begin{theorem}
\label{th:aligned_gradient_variance}
Assume $C$ is chosen such that the bias induced by clipping is the same in vanilla and aligned DPVI. Then for any fixed batch,
\begin{equation}
    \Var{\draw,\dpdraw}{ \tilde{g}_s^{aligned} } \leq \Var{\draw,\dpdraw}{ \tilde{g}_s^{vanilla} },
\end{equation}
where $\draw$ is the random variable of the MC approximation to the ELBO and $\dpdraw$ that of the DP perturbation.
\end{theorem}

\paragraph{Aligned Natural Gradients}

Finally we also consider a combination of natural gradients and aligning, to enable the benefits of natural gradients for convergence while simultaneously removing the need to consider the gradient of $\qs$ for DP clipping and perturbation. The full procedure is given in Algorithm~\ref{alg:ng_aligned}. We use $\mathcal{I}_m, \mathcal{I}_s$ to refer to the blocks of $\mathcal{I}$ corresponding to the gradient components.

\begin{algorithm}[h]
    \begin{algorithmic}
    \State $\params \gets \qm + \draw T(\qs)$ where $\draw \sim \mathcal{N}(0,I)$
    \State $\gqmx^{nat} \gets \mathcal{I}_m^{-1} \gm \elbo(q)$ for $\bx \in \data$ %\Comment{Compute per-example natural gradients for $\qm$}
    \State $\gamma_{\bx} \gets \min(1, \nicefrac{C}{||\gqmx^{nat}||})$ for $\bx \in \data$ \Comment{clipping factor} %\Comment{Compute the clipping multiplier}
    \State $\dpgqm^{nat} \gets \sum_{\bx \in \data} \gamma_{\bx} \gqmx + \sigma_{DP} C \dpdraw$, where $\dpdraw \sim \mathcal{N}(0,I)$ %\Comment{Get DP release for $\gqm^{nat}$}
    \State $\dpgqs^{nat,aligned} = \mathcal{I}_s^{-1} (\draw T'(\qs) \mathcal{I}_m \dpgqm^{nat} + \gs H(q)).$ %\Comment{Get DP aligned $\gqs^{nat}$ via post-processing}
    \end{algorithmic}
    \caption{The aligned natural gradient procedure \label{alg:ng_aligned}}
\end{algorithm}

\subsubsection{Extending to full-rank covariance matrices}
So far we have only considered a diagonal Gaussian as the variational posterior. Due to the low dimensionality of the variational parameters, this approach is computationally effective and often applied in practice, but it has limitations: It cannot capture correlations among different model parameters and, more importantly, it will underestimate the marginal variances of the parameters when the true covariance structure is non-diagonal. For those reasons, a full-rank covariance approximation would be favored to correctly capture the uncertainty of the parameters. 

However, learning the full-rank covariance approximation results in a quadratic (in the number of dimensions $d$) expansion of the number of learnable parameters. This not only increases computational costs but also implies less accurate learning of the parameters under DP, as the available privacy budget has to be spread over more parameters. Fortunately, the aligning procedure can be extended to full-rank Gaussian approximations as well, which allows us to alleviate the issue of increased sensitivity. The proof is very similar to the diagonal case. Instead of the parameters $\qs$ corresponding to marginal standard deviations, we now consider a parameter vector $\qa \in \mathbb{R}^{\frac{d(d+1)}{2}}$ and a transformation function $T: \mathbb{R}^{\frac{d(d+1)}{2}} \rightarrow \mathbb{R}^{d \times d}$ such that $T(\qa)$ corresponds to the Cholesky factor of the posterior covariance. That is, $T$ must guarantee that $T(\qa)$ is a lower triangular with positive entries along its diagonal, which will require similar transformations as in the purely diagonal covariance case discussed previously. Now, the reparametrisation step in \eqref{eq:reparam} becomes
\begin{align}
    \theta := \qm + T(\qa) \draw,
    \label{eq:reparam_full}
\end{align}
and the gradient w.r.t $\qa$ can be written as 
\begin{align}
    \gqa = J_a(T(\qa) \draw) \gqm + \nabla_{\qa} H(q),
\end{align}
where $J_a$ denotes the Jacobian of $T$ w.r.t $\qa$. Therefore, the gradient $\gqa$ can again be written as a data-independent transformation of $\gqm$. Thus under the post-processing immunity of DP, we can get the DP gradients for $\qa$ from DP versions of $\gqm$ without suffering the quadratic increase of the size of the input to underlying the Gaussian mechanism present in vanilla DPVI.

%%%%%%%%%%%%%%%%%%%%%%%%%%%%%%%%%%%%%%%%%%%%%%%%%%%%%%%%%%%%%%

\subsection{DPVI samples noisy parameter estimates}

As other applications of DP-SGD, vanilla DPVI does not take into account the uncertainty that the DP mechanism introduces. Instead, after a (finite) number of iterations, the values found in the last iteration are usually treated as the true variational parameters. We argue that treating them this way can lead to severe errors in accuracy because these values are merely a noisy estimate of the optimal values. Annealing the learning rate does not help: Depending on when the annealing is started, the distance to an optimum can still be large due to the random walk prior to annealing.

\citet{MandtHB17} investigated a random walk behaviour around the optimum for regular (non-DP) SGD induced by the noise arising from subsampling. They assume that near the optimum $\qparams^*$ the loss function is well approximated by a quadratic approximation $L(\qparams) \approx \frac{1}{2} (\qparams-\qparams^*)^T \boldsymbol{A} (\qparams-\qparams^*)$, and show that the stochastic process around the optimum can be characterised as an Ornstein-Uhlenbeck (OU) process
\begin{align}
    d\qparams(t) = -\alpha \boldsymbol{A} (\qparams(t) - \qparams^*)dt + \frac{1}{\sqrt{S}} \alpha \boldsymbol{B} dW(t),
    \label{ou_process}
\end{align}
where $W(t)$ is a Wiener process, $\alpha$ is the step size of the SGD (assumed constant), $S$ is the size of the subsampled data and $\boldsymbol{B}$ the Cholesky decomposition of the covariance matrix $\boldsymbol{Z}$ of the noise due to the subsampling. Directly adapting this analysis, we suggest that under the same regularity assumptions, DP-SGD still is an OU process. The principle of the proof is straightforward: DP-SGD adds an additional Gaussian noise component, allowing us to add the (diagonal) covariance matrix of the DP noise to $\boldsymbol{Z}$ and obtain a $\hat{\boldsymbol{B}}$ such that
\begin{align}
    \boldsymbol{Z} + \sigma_{DP}^2 I = \hat{\boldsymbol{B}}\hat{\boldsymbol{B}}^T.
\end{align}
The more detailed proof can be found in Appendix~\ref{app:DPVI_OU}. This insight, combined with the fact that the privacy guarantees of DPVI hold for all released intermediate parameter values, allows us to make the following two suggestions to improve the parameter estimates of DPVI.

%%%%%%%%%%%%%%%%%%%%%%%%%%%%%%%%%%%%%%%%%%%%%%%%%%%%%%%%%%%%%%

% \subsubsection{Iterate averaging}
\paragraph{Iterate averaging to reduce noise in parameter estimate}

In order to reduce noise in our learned variational parameters, we apply iterate averaging \citep{PolyakBJ92} and average the parameter traces, i.e., the sequence of parameters during optimisation, over the last $\Tbo$ iterates for which we assume the trace has converged. As the OU process in Equation \eqref{ou_process} is symmetric around the optimum, the mean of the trace is an unbiased estimator of  $\qparams^*$. Compared to using the final iterate of the chain (also an unbiased estimator of $\qparams^*$), the averaged trace reduces the variance of the estimator by up to a factor of $\Tbo^{-1}$.

%%%%%%%%%%%%%%%%%%%%%%%%%%%%%%%%%%%%%%%%%%%%%%%%%%%%%%%%%%%%%%

% \subsubsection{Estimating the increased variance due to DP \label{sec:var_est}}
\paragraph{Estimating the increased variance due to DP \label{sec:var_est}}
Finally, since our posterior approximation is Gaussian and the stationary distribution of the OU is Gaussian as well, we can add the variance of the averaged traces to the variances of our posterior to absorb the remaining uncertainty due to the inference process, and recover a noise-aware posterior approximation.

Now the remaining problem is to determine $\Tbo$, the length of the trace where the parameters have converged. For this we suggest a simple convergence check based on linear regression: For each of the traces, we fit linear regression models over different candidate $\Tbo$. 
The regressor $\bX_{linreg}$ is set to interval $[0,1]$ split to $\Tbo$ points in ascending order. The responses $y$ are set to the corresponding parameter values in the trace, e.g. $\mathbf{y} = \{{\qm}^{(t)}\}_{t=T-T_{burn-out}}$.
If the linear regression model has a sufficiently small slope coefficient, we consider the trace as converged and pick the longest $\Tbo$ for which this is the case.

%%%%%%%%%%%%%%%%%%%%%%%%%%%%%%%%%%%%%%%%%%%%%%%%%%%%%%%%%%%%%%

\section{Experiments \label{sec:exp}}

\begin{figure*}[tbh]
    \centering
    \begin{subfigure}[t!]{0.48\textwidth}
        \centering
        \includegraphics[width=\textwidth]{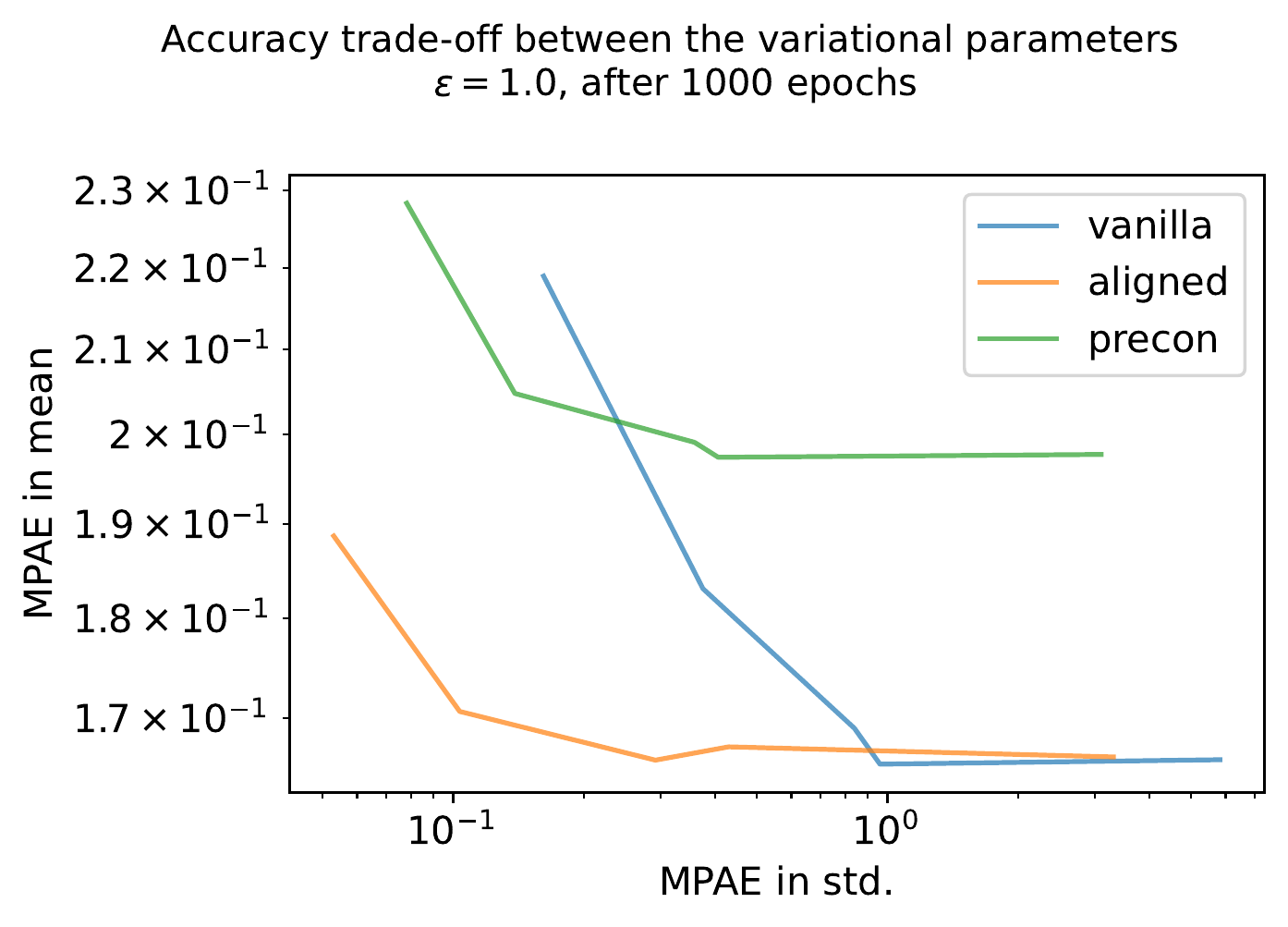}
        \caption{\label{fig:tradeoff_ukb}}
    \end{subfigure}
    \hfill
    \begin{subfigure}[t!]{0.48\textwidth}
        \centering
        \includegraphics[width=\textwidth]{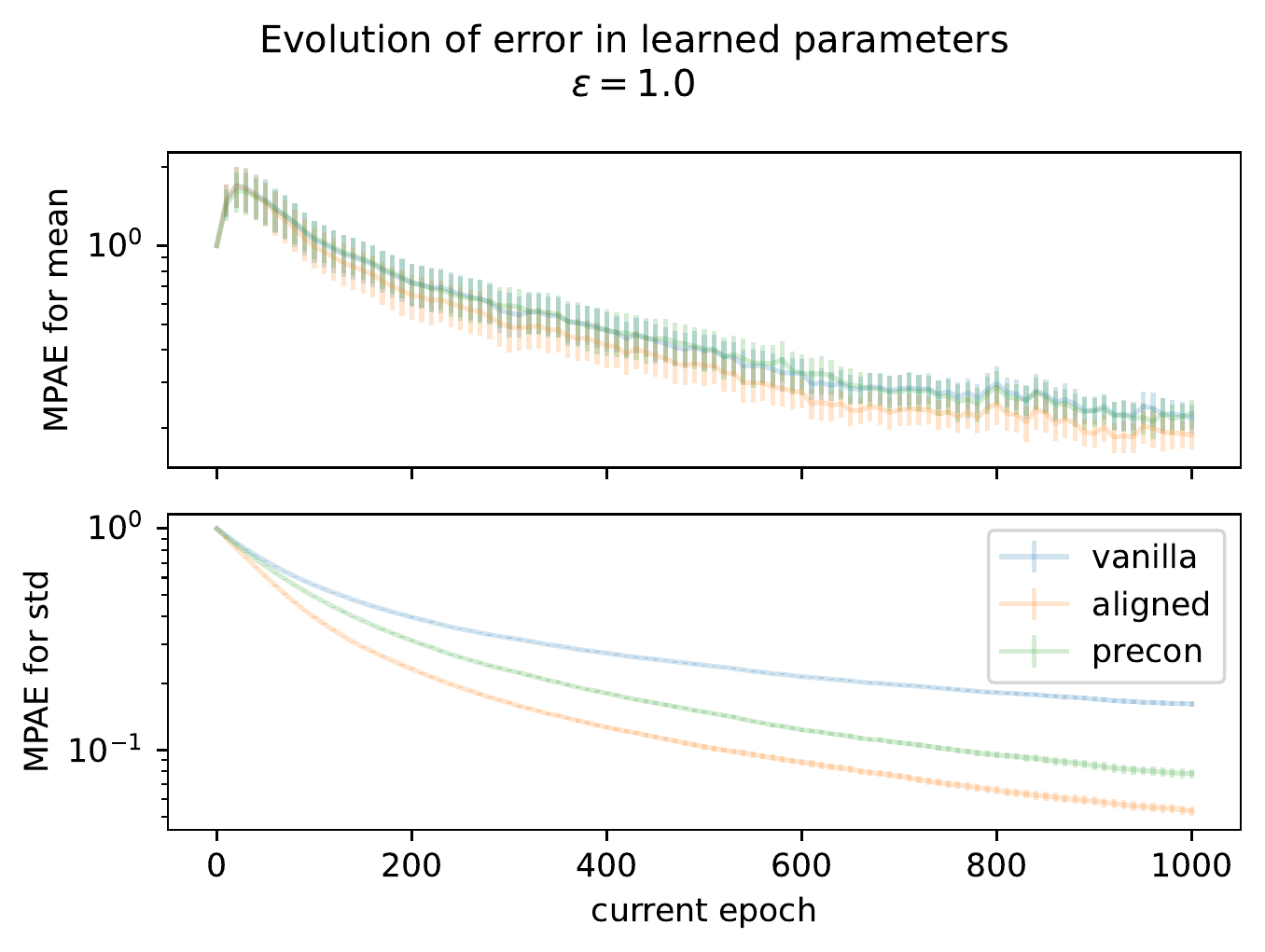}
        \caption{\label{fig:init_test_ukb_trace_1.0}}
    \end{subfigure}
    \caption{(a) Aligned DPVI makes the smallest error in learning the variational parameters across all different initial values for $\qsig$, implying it is the most robust. (b) Aligned DPVI converges faster than the other variants while also having less deviation across the repeats (all initialised at $\qsig=1$). Both subfigures show averaged MPAE for vanilla, aligned and preconditioned DPVI with error bars in (b) indicating standard error over repeats.}
\end{figure*}

We experimentally test our methods for two different tasks using mean-field approximation with real data: learning a probabilistic generative model for private data sharing and learning a logistic regression model. We also experimentally explore aligned DPVI with full-rank Gaussian approximation using simulated data.

%%%%%%%%%%%%%%%%%%%%%%%%%%%%%%%%%%%%%%%%%%%%%%%%%%%%%%%%%%%%%%

\subsection{Implementation details}

We implemented the different variants for DPVI introduced in Sec.~\ref{sec:dpvi_variants} using the d3p package \citep{PredigerLKH22} for the NumPyro probabilistic programming framework \citep{PhanPJ19, Bingham+19}. To compute the privacy cost in our experiments, we use the Fourier accountant method \citep{KoskelaJPH21}. The hyperparameters used in our experiments are discussed in the Appendix \ref{app:hyperparameters}. We use the softplus function as our transformation $T$ in all experiments. The code for reproducing the experiments can be found at \url{https://github.com/DPBayes/dpvim-experiments}.

In order to asses learning over multiple parameters which converge to different values, and over repeated runs with different initial values, we define a \emph{mean proportional absolute error (MPAE)}: Let $\qparams^{(t)} \in \mathbb{R}^D$ be the parameter vector at iteration $t$ and $\qparams^*$ be the parameter vector at the optimum.\footnote{Since the optimal value $\qparams^*$ is typically unknown, we instead use the results of classical non-DP variational inference in its place in practice.} We measure the MPAE at iteration $t$ as
\begin{equation}
    \label{eq:MPAE}
    \text{MPAE}(\qparams^{(t)}) = \frac{1}{D} \sum_{d=1}^D \frac{|\qparams_d^{(t)} - \qparams_d^*|}{|\qparams_d^{(0)} - \qparams_d^*|}.
\end{equation}
An MPAE value of $0$ indicates perfect recovery of the optimum, a value of $1$ suggests that the parameters on average did not move away from their initialisation.

%%%%%%%%%%%%%%%%%%%%%%%%%%%%%%%%%%%%%%%%%%%%%%%%%%%%%%%%%%%%

\subsection{Using DPVI to learn a generative model}

Recently, \citet{JalkoLHTHK21} suggested using DPVI to learn a probabilistic generative model for differentially private data sharing. Note that in this application it is especially crucial to learn the posterior variances well to faithfully reproduce the uncertainty in the original data in the synthetic data set. 

A recent study by \citet{Niedzwiedz+20} on personal health data from the United Kingdom Biobank (UKB) \citep{Sudlow+15} studied how socio-economic factors affect an individual's risk of catching the SARS-CoV-2 virus. We aim to produce synthetic data, using DPVI to learn the generative model, from which we can draw similar discoveries.

Following \citet{Niedzwiedz+20}, we consider a subset of UKB data which comprises of \numprint{58261} individuals with $d=7$ discrete (categorical) features. We split the features into a set of explanatory variables and a response variable indicating whether the individual was infected by SARS-CoV-2.

We place a mixture model for the explanatory variables $\bX$, and a Poisson regression model mapping the explanatory variables to the responses $\by$, using $\params_{\bX}$, $\params_{\by}$ and $\boldsymbol{\pi}$ to designate the model parameters:
\begin{align}
    &p(\bX \mid \params_{\bX}, \bpi) = 
            \sum_{k=1}^{K} \pi_k \prod_{j=1}^d \text{Categorical}(\bX_j \mid \params_{\bX}^{(k)}) \\
    &p(\by \mid \bX, \params_{\by}) = \text{Poisson}(\by \mid \exp(\bX \params_{\by})).
\end{align}
In our experiments, we set the number of mixture components $K=16$ which was chosen based on internal tests. Priors for the model parameters are specified in Appendix~\ref{app:ukb_model_priors}.

\paragraph{Aligned DPVI is more robust to initialisation}
We first demonstrate that aligned DPVI improves robustness to initialisation over vanilla and preconditioned DPVI. To do so we fix a privacy budget of $\varepsilon=1$ and the number of passes over the entire data set, i.e., \emph{epochs}, to $1000$ and vary the initial value of $\qs$ such that $\qsig$ is between $0.01$ and $1$. We perform 10 repetitions with different random seeds over which we keep the initialisation of $\qs$ fixed but initialise $\qm$ randomly. We compute the MPAE over the parameters of the Poisson regression part in the model, which corresponds directly to the downstream prediction task we are ultimately interested in. 

Figure~\ref{fig:tradeoff_ukb} shows the trade-off different variants of DPVI make between the MPAE in variational means ($\qm$) and stds ($\qs$) averaged over the 10 repetitions for the different initial values of $\qs$. We observe that the aligned variant is able to achieve small errors in $\qm$ and $\qs$ simultaneously while the alternatives cannot. To see how the MPAEs for $\qm$ and $\qs$ behave individually w.r.t. the initial value for $\qs$ refer to Appendix \ref{app:more_robustness_results}. The natural gradient as well as the aligned natural gradient method performed poorly in this experiment and we report results for them in the appendix as well.

\begin{figure}[tb]
    \hspace*{\fill}%
    \begin{minipage}[t]{.48\textwidth}
        \centering
        \vspace{0pt}
        \includegraphics[width=\columnwidth]{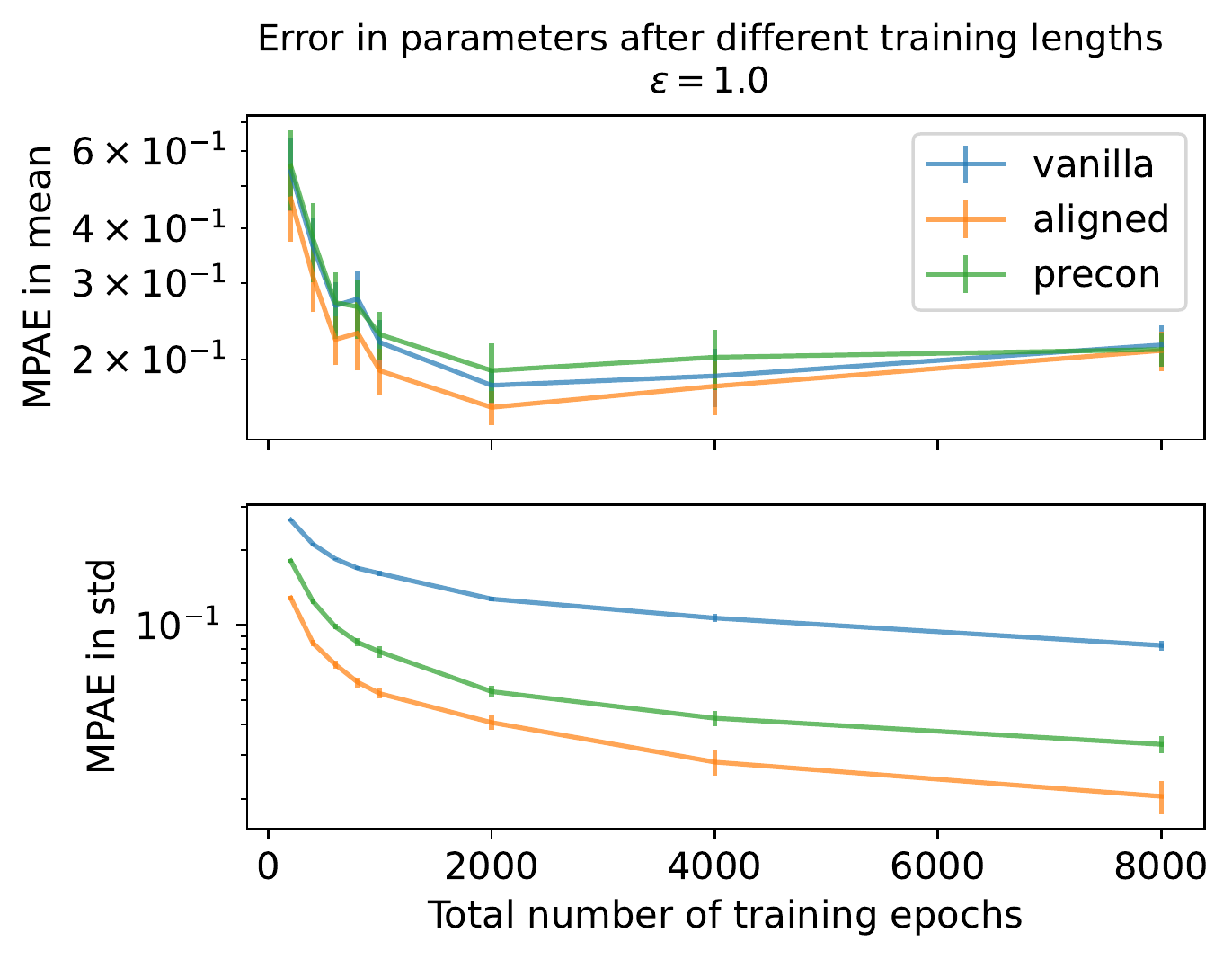}
        \caption{The aligned variant remains the most accurate method, even if we run the algorithm for longer. Initial $\qsig = 1$. \label{fig:num_epochs_test_ukb}}   
    \end{minipage}
    \hfill
    \begin{minipage}[t]{.48\textwidth}
        \centering
        \vspace{0pt}
        \includegraphics[trim={0 0 0 5mm},clip,width=\columnwidth]{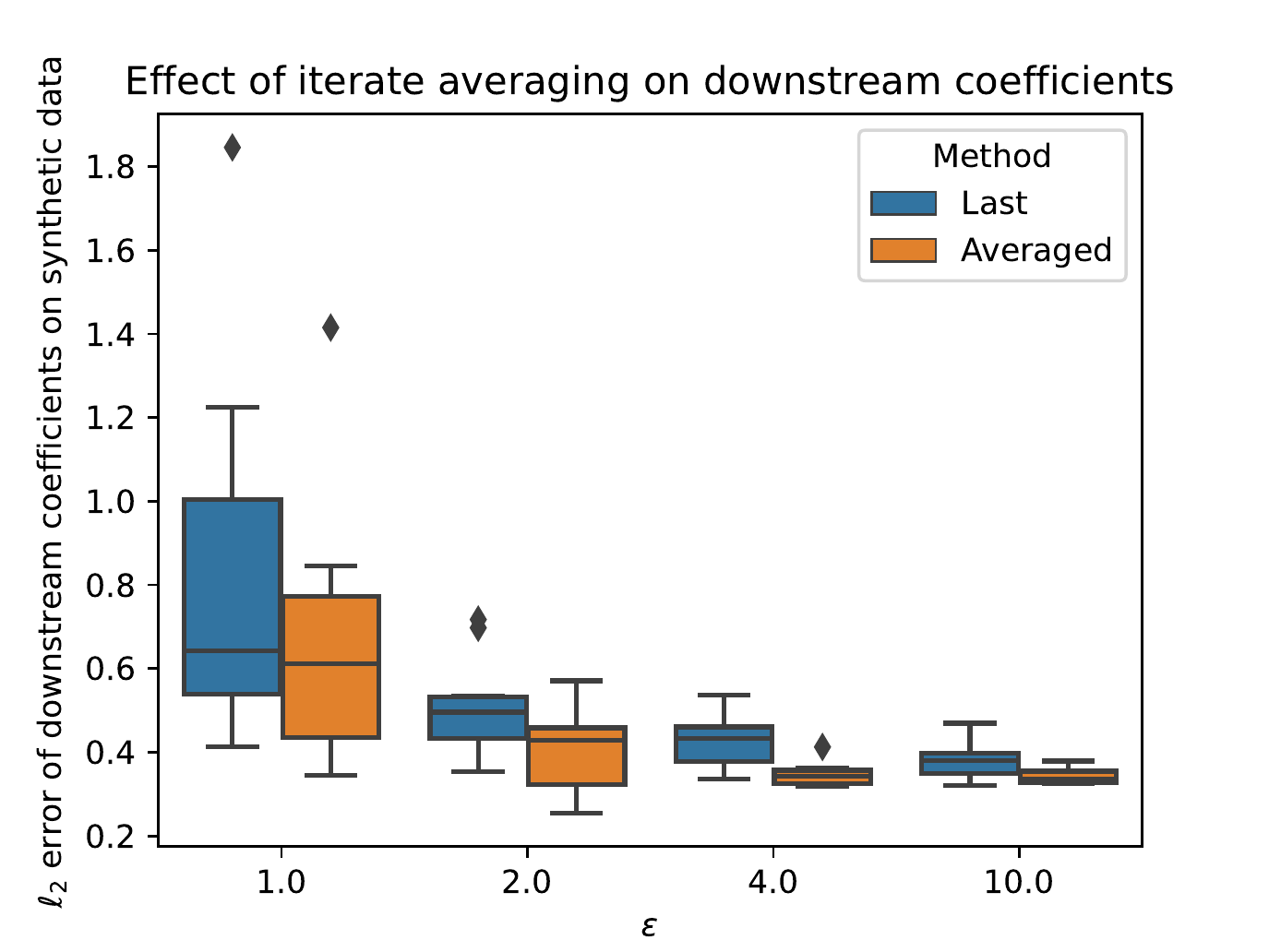}
        \caption{RMSE of parameters found in downstream analysis when doing iterate averaging with different $\Tbo$. Iterative averaging can reduce the error and significantly reduce variance of error compared to using only the last iterate. $\varepsilon=1$, initial $\qsig=1$, $4000$ epochs of aligned DPVI.}
        \label{fig:iter_avg_ukb_downstream}
    \end{minipage}
    \hspace*{\fill}%
\end{figure}

\paragraph{Longer runs do not help vanilla DPVI}

Figure~\ref{fig:init_test_ukb_trace_1.0} suggests that vanilla DPVI has not converged in terms of MPAE, in the allotted number of iterations for an initial $\qsig=1$. An obvious solution then seems to be to run the inference for longer. We now fix the initialisation of $\qsig$ to $1$, which designates the least relative scaling of gradients at the beginning of training and thus a best-case scenario for vanilla DPVI. We vary the number of training epochs from $200$ to $8000$ while always keeping the privacy budget fixed at $\varepsilon=1$. Since longer runs require more accesses to the data, the DP perturbation scale increases with the number of iterations. As before, we repeat 10 inference runs for each parameter choice.

Figure~\ref{fig:num_epochs_test_ukb} shows the final MPAE over all 10 repetitions and all parameters in the Poisson regression part of the model for the different numbers of total epochs.\footnote{Note that it is NOT showing the evolution of the error over a single training of 8000 epochs.} The upper panel shows that with an increasing number of epochs, the difference in MPAE of variational means between vanilla and aligned DPVI vanishes. However, the lower panel shows clearly that even in a long training regime, vanilla DPVI still does not converge in variational variances and is consistently beaten by our aligned variant.

\paragraph{Iterate averaging increases robustness of downstream task}

Next, we test the iterate averaging of noisy parameters traces for the generative model. We use the linear regression technique discussed in Sec.~\ref{sec:var_est}, individually for each parameter, to determine the length of the trace to average. We then use synthetic data from the generative model to learn the Poisson regression model used by \citet{Niedzwiedz+20} and compare the regression coefficients against those obtained from the original data. Further details on the downstream analysis setup are given in Appendix \ref{app:downstream_details}. Figure \ref{fig:iter_avg_ukb_downstream} shows that the results from iterate averaged model are less noisy compared to just using the last iterate as the true parameters. However, the approach appears to be somewhat unstable: Changing the initial values of $\qsig$ to $0.1$ causes the variance of error for $\varepsilon=1.$ to increase over the non-averaged case. This is likely due to the simple linear regression heuristic we used in this experiment not detecting convergence correctly in this case.

%%%%%%%%%%%%%%%%%%%%%%%%%%%%%%%%%%%%%%%%%%%%%%%%%%%%%%%%%%%%%%

\begin{figure*}[tb]
    \begin{subfigure}[b]{.52\textwidth}
        \centering
        \vspace{0pt}
        \includegraphics[width=\columnwidth]{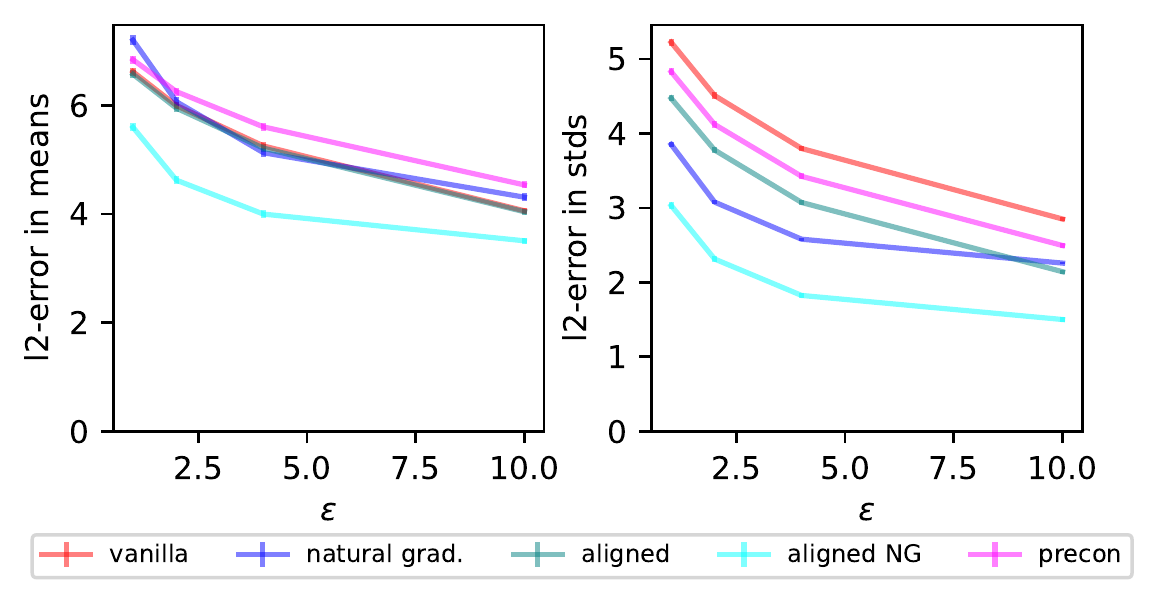}
        \caption{ \label{fig:adult_l2}}
    \end{subfigure}
    \hspace*{\fill}%
    \begin{subfigure}[b]{.45\textwidth}
        \centering
        \vspace{0pt}
        \includegraphics[width=\columnwidth]{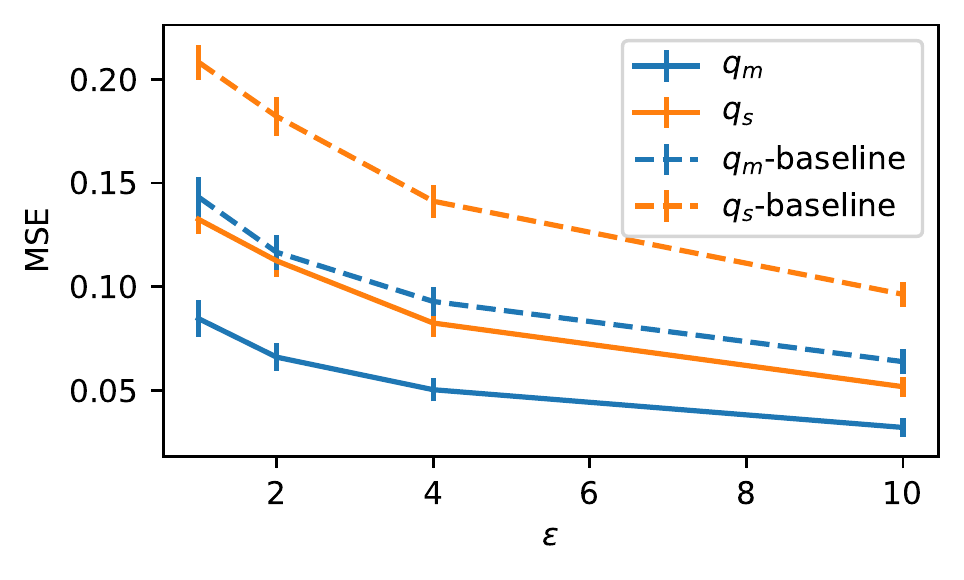}
        \caption{\label{fig:std_estimation}}
    \end{subfigure}
    \caption{(a) The aligned natural gradient method is closest to the non-private variational parameters. Error is computed as a mean $\ell_2$-norm against non-private baseline over 20 repeats. Error bars show the standard error of the mean. (b) The standard deviation inferred from the converged traces (determined by the linear regression method) is close to that computed over last iterates across different repeats. Lines show the average MSE, error bars show the standard deviation across repeats. The baselines show the mean squared norm of the standard deviation estimated from repeated runs, corresponding to the MSE for not estimating DP-induced noise.}
\end{figure*}

\subsection{Experiments on Adult data set}

As the UK Biobank data is access-restricted, we further demonstrate our methods on the publicly available Adult data set from the UCI machine learning repository \citep{DuaG19}, which contains \numprint{30162} training records. We learn a logistic regression model, classifying whether the income feature of the data exceeds $\$50$k based on all other features.

\paragraph{Aligned natural gradient outperforms the other variants}
We compare our private logistic regression coefficients to the ones obtained using privacy-agnostic VI. We also test the aligned natural gradient and the natural gradient variants for this data. Figure \ref{fig:adult_l2} shows the $\ell_2$-norm between variational parameters learned with and without DP. From this figure we see that the aligned natural gradient method clearly outperforms all the other variants in this setting. Additionally, we clearly see again that vanilla DPVI learns the stds poorly, and also that the natural gradient variant reverses the problem compared to vanilla and struggles in learning the variational means as suggested in Section \ref{sec:dpvi_variants}.

\paragraph{DP noise can be inferred from the (converged) traces}

We test how well we can recover the DP noise effect from the parameter traces. We limit the test to the coefficients that have converged according to the linear regression test described in Section \ref{sec:var_est}. Based on our internal tests, we chose a slope of $0.05$ as the threshold for convergence. We compare the standard deviation of the converged parameter trace to an estimate of the DP-induced noise estimated by the standard deviation of the last iterates over 50 repeats in terms of mean squared error over different parameter sites. 

Figure \ref{fig:std_estimation} shows that the noise std estimated from the converged traces is close to the noise std we have across the last iterates of multiple independent repeats.

\subsection{Experiments with full-rank covariance}
\label{sec:exp_fullrank}
As a final experiment, we investigate aligned gradients for a full-rank Gaussian posterior approximation. We perform Bayesian linear regression over a simulated data set where we can control the number of feature dimensions as well as the strength of correlations. We control the latter by setting the rate of nonzero off-diagonal entries in the covariance matrix for simulated data points. Further details of the experimental setup can be found in Appendix~\ref{app:full_rank_experiment}. 

Figure~\ref{fig:full_rank_experiment} confirms that aligned gradients improve the average predictive log-likelihood of the posterior approximation over a held-out test set. This is true even when the data is not strongly correlated (panels in the right column), as the large increase in parameters over which vanilla DPVI has to split the privacy budget negatively impacts the learning.

\begin{figure}[tb]
    \centering
    \includegraphics[width=.48\columnwidth]{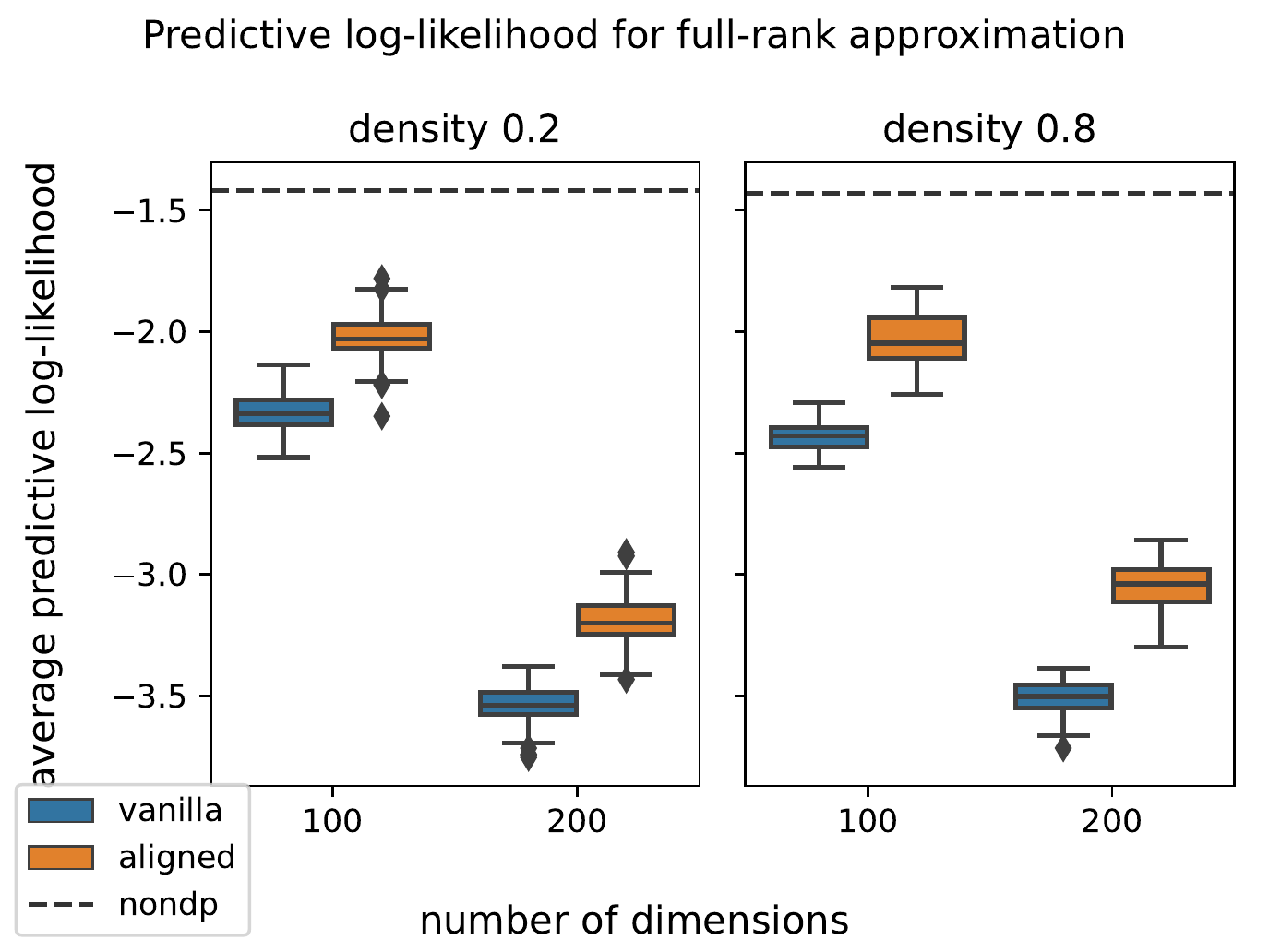}
    \caption{DPVI with aligned gradients achieves better predictive log-likelihood for a full-rank approximation than vanilla DPVI. Higher density means more nonzero entries in data covariance. $\varepsilon=1$. 50 repetitions.}
    \label{fig:full_rank_experiment}
\end{figure}

%%%%%%%%%%%%%%%%%%%%%%%%%%%%%%%%%%%%%%%%%%%%%%%%%%%%%%%%%%%%%%

\section{Discussion \label{sec:conc}}

In this paper we introduced the aligned gradient solution for the specific task of learning a mean-field variational posterior. The technique should be applicable also in other tasks where gradients with respect to different parameters depend on data through a common term. Detection of such cases could be even automated by inspecting how the data enters the computational graph of the task. This would be an interesting future direction.

A limitation of DP in general is that it guarantees indistinguishability among the individuals in the data set by aiming to preserve more common characteristics of the data. Therefore the utility of a DP algorithm might be worse for individuals from less common groups.

The poor performance of aligned natural gradients in the UKB example might be due to bad hyperparameter choices. While we performed some hyperparameter tuning without success, a more comprehensive search would be needed to fully assess the performance of these methods.

The literature on MCMC holds many existing diagnostics for the converge of chains, such as the (split) R-hat estimator \citep{GelmanARD92, Vehtari+21}, that could be used to test the convergence of the parameter traces as well. Tuning these methods to work well in diagnosing the converge of the parameter trace would require extensive testing which we leave for future work.

%%%%%%%%%%%%%%%%%%%%%%%%%%%%%%%%%%%%%%%%%%%%%%%%%%%%%%%%%%%%%%
\section*{Acknowledgements}
This work was supported by the Academy of Finland (Flagship programme: Finnish Center for Artificial Intelligence, FCAI; and grants 325572, 325573), the Strategic Research Council at the Academy of Finland (Grant 336032), UKRI Turing AI World-Leading Researcher Fellowship, EP/W002973/1 as well as the European Union (Project 101070617). Views and opinions expressed are however those of the author(s) only and do not necessarily reflect those of the European Union or the European Commission. Neither the European Union nor the granting authority can be held responsible for them. The authors also acknowledge the computational resources provided by the Aalto Science-IT project.

This research has been conducted using the UK Biobank Resource under Project Number 65101.
%%%%%%%%%%%%%%%%%%%%%%%%%%%%%%%%%%%%%%%%%%%%%%%%%%%%%%%%%%%%%%

\printbibliography[segment=0]

%%%%%%%%%%%%%%%%%%%%%%%%%%%%%%%%%%%%%%%%%%%%%%%%%%%%%%%%%%%%%%

%%%%%%%%%%%%%%%%%%%%%%%%%%%%%%%%%%%%%%%%%%%%%%%%%%%%%%%%%%%%
%Appendix
%%%%%%%%%%%%%%%%%%%%%%%%%%%%%%%%%%%%%%%%%%%%%%%%%%%%%%%%%%%%

\onecolumn
\newpage
\appendix
\appendixpage
\counterwithin{figure}{section}
\counterwithin{table}{section}
\counterwithin{algorithm}{section}
\numberwithin{equation}{section}

\newrefsegment %% <== increases the segment number (0 by default)

\section{Intuitive reasoning why larger noise would slow convergence}
\label{app:larger_noise_slows_convergence}
We consider only a single optimisation step in a single dimension for simplicity.
Assume we are at $\theta^{(t)}$ and have noisy gradient
\begin{equation}
    g^{(t)} = \nabla \elbo(\theta^{(t)}) + \eta, \eta \sim \mathcal{N}(0, \sigma^2)
\end{equation}
for some perturbation scale $\sigma$. We update the parameter as
\begin{equation}
    g^{(t+1)} = \theta^{(t)} - \alpha g^{(t)}
\end{equation}
with learning rate $\alpha$.

In order to get closer to the optimum, we want $sign(g^{(t)}) = sign(\nabla \elbo(\theta^{(t)}))$. Assume wlog that $\nabla \elbo(\theta^{(t)}) > 0$, then
\begin{align}
    \Pr\left[ sign(g^{(t)}) = sign(\nabla \elbo(\theta^{(t)})) \right]
    &= \Pr\left[ \nabla \elbo(\theta^{(t)}) + \eta > 0 \right] \\
    &= \Pr\left[ \eta > -\nabla \elbo(\theta^{(t)}) \right] \\
    &= 1 - \Pr\left[ \eta \leq -\nabla \elbo(\theta^{(t)}) \right] \\
    &= 1 - \Phi(-\nabla \elbo(\theta^{(t)})) \\
    &= \Phi(\nabla \elbo(\theta^{(t)})) \\
    &= \frac{1}{2}\left(1 + \erf\left(\frac{\nabla \elbo(\theta^{(t)})}{\sigma \sqrt{2}}\right) \right).
\end{align}

$\erf(\cdot)$ is a monotonically increasing function, so we see from the above that the probability of progressing towards the optimum decreases with decreasing $\frac{\nabla \elbo(\theta^{(t)})}{\sigma}$. I.e., for a fixed gradient, larger variance $\sigma^2$ will decrease the probability of progressing towards the optimum in each step.

\section{Proof of Proposition \ref{thm:gs_from_gm}}
\label{app:proposition_argument}

The proposition follows quite easily from $\gm \params(\draw; \qm, \qs) = 1$ (from Eq.~\eqref{eq:reparam} inserted into Eq.~\eqref{eq:mc_m_grad}:
\begin{align}
    \gqm &= \gm \elbo(q) = \gparams \log p(\data, \params(\draw; \qm, \qs)) 
\end{align}
Inserting this in turn into Eq.~\ref{eq:mc_s_grad} yields
\begin{align}
    \gqs &= \gs \log p(\data, \params(\draw; \qm, \qs)) + \gs H(q) \\
    &= \draw \gs T(\qs) \gparams \log p(\data, \params) + \gs H(q) \\
    &= \draw T'(\qs) \gqm  + \gs H(q).
\end{align}

\section{Proof of $T'(\qs) \leq T(\qs)$ for softplus and exponential function}
\label{app:T_proof}
\paragraph{$T(\qs) = \text{softplus}(\qs)$}
Consider we transform $\qs$ in positive real numbers using the softplus function:
\begin{align}
    T(\qs) = \log (1 + \exp(\qs)).
\end{align}
First, we make the following observation which connects the softplus to the sigmoid function
\begin{align}
    T(\qs) &= -\log \left( \frac{1}{1+\exp(\qs)} \right) \nonumber\\
    &= -\log \left( 1 - \frac{1}{1+\exp(-\qs)} \right) \nonumber \\
    &= -\log \left( 1 - \sigma(\qs) \right),
\end{align}
where $\sigma$ denotes the sigmoid function. We then get $T'(\qs) = \sigma(\qs)$. It is easy to see that $\log(x) \leq x-1$ and hence 
\begin{align}
    T(\qs) &= -\log \left( 1 - \sigma(\qs) \right) \nonumber \\
    &\geq 1-(1-\sigma(\qs)) = \sigma(\qs) = T'(\qs).
\end{align}
We have therefore shown that $T(\qs) \geq T'(\qs)\, \forall \qs \in \mathbb{R}$.

\paragraph{$T(\qs) = \exp(\qs)$}
For $T(\qs) = \exp(\qs)$ the proof follows immediately from the fact that $T'(\qs) = \exp(\qs) = T(\qs)$.

\section{Proof of Theorem~\ref{th:aligned_gradient_variance}: Variance in aligned scale gradients is smaller}
\label{app:proof_aligned_gradients_smaller_scale}

Per Eq.~\ref{eq:mc_s_grad_as_theta} we get for the perturbed gradient for $\qs$ in vanilla DPVI
\begin{align}
    \dpgqs &= \alpha \gs \elbo(q) + \dpdraw \sigma_{DP} C \\
                &= \alpha \left(\draw T'(\qs) \gm \elbo(q) + \gs H(q)\right) + \dpdraw \sigma_{DP} C
\end{align}
where $\dpdraw \sim \mathcal{N}(0, 1)$ is the DP perturbation draw and $\draw \sim \mathcal{N}(0,1)$ the draw for the reparameterisation approach. $\alpha$ is the clipping multiplier, i.e.,
\begin{align}
    \alpha = \min\left(1, \nicefrac{C}{||g||}\right),
\end{align}
where $g = (\gm \elbo(q), \gs\elbo(q))^T$ is the combined gradient.

Assuming fixed batch and parameters, the variance of $\dpgqs$ with respect to $\dpdraw$ and $\draw$ is
\begin{align}
    &\Var{\draw,\dpdraw} { \alpha \left( \draw T'(\qs)  \gm\elbo(q) + \gs H(q) \right) + \dpdraw \sigma_{DP} C } \\
    &= \Var{\draw}{ \Exp{\dpdraw|\draw}{\alpha \left( \draw T'(\qs) \gm \elbo(q) + \gs H(q) \right) + \dpdraw \sigma_{DP} C} }\\
    &\hspace{1em} + \Exp{\draw}{\Var{\dpdraw|\draw} { \alpha \left(\draw T'(\qs) \gm \elbo(q) + \gs H(q) \right) + \dpdraw \sigma_{DP} C } } \\
    &= \Var{\draw}{ \alpha \left(\draw T'(\qs) \gm \elbo(q) + \gs H(q) \right) } + \Exp{\draw}{\sigma_{DP}^2 C^2 } \\
    &= T'(\qs)^2 \Var{\draw}{ \draw \alpha \gm \elbo(q) } + \sigma_{DP}^2 C^2.
\end{align}

The perturbed aligned gradient is obtained as:
\begin{align}
    \tilde{g}_s^{aligned} &= \draw T'(\qs) \tilde{g}_m + \gs H(q) \\
    &= \draw T'(\qs) \left(\alpha \gm \elbo(q) + \dpdraw \sigma_{DP} C\right) + \gs H(q) \\
\end{align}
and its variance is
\begin{align}
    &\Var{\draw,\dpdraw}{ \draw T'(\qs) \left(\alpha \gm \elbo(q) + \dpdraw \sigma_{DP} C\right) + \gs H(q) } \\
    &= \Var{\draw}{ \Exp{\dpdraw|\draw}{\draw T'(\qs) \left(\alpha \gm \elbo(q) + \dpdraw \sigma_{DP} C\right) + \gs H(q)} }\\
    &\hspace{1em} + \Exp{\draw}{\Var{\dpdraw|\draw}{ \draw T'(\qs) \left(\alpha \gm \elbo(q) + \dpdraw \sigma_{DP} C\right) + \gs H(q) } } \\
    &= \Var{\draw}{ \draw T'(\qs) \alpha \gm \elbo(q) + \gs H(q) } + \Exp{\draw}{ \draw^2 T'(\qs)^2 \sigma_{DP}^2 C^2 } \\
    &= T'(\qs)^2 \Var{\draw}{ \draw  \alpha \gm \elbo(q) } + T'(\qs)^2 \sigma_{DP}^2 C^2 \underbrace{\Exp{\draw}{ \draw^2 }}_{=1}
\end{align}

These variances differ only in the additional scaling of the DP perturbation term and it follows directly that $\Var{\draw,\dpdraw}{ \tilde{g}_s } \geq \Var{\draw,\dpdraw}{ \tilde{g}_s^{aligned} }$ iff $T'(\qs)^2 \leq 1$.

Note that due to the clipping, $\Var{\draw}{ \draw \alpha \gm \elbo(q) } \leq \Var{\draw}{ \draw C } = C^2$, which suggests that the first term in the total gradient variant expressions will not exceed the term due to DP perturbation (again, for small $T'(s)'$).

The above assumes that $C$ stays constant between both variants, which is a reasonable assumption for small $T'(s)$. However we now argue that for larger values of $T'(s)$ different clipping thresholds should be chosen for both variants and then the assertion about the variances holds for all $T'(s)$.

We obtain the norm of the combined gradient as
\begin{align}
    ||g||_2 &= \sqrt{||g_m||_2^2 + ||g_s||_2^2} \geq \sqrt{1 + T'(s)^2\draw^2} ||g_m||_2,
\end{align}
where the last inequation omits the entropy term $\gs H(q) = \nicefrac{T'(s)'}{T(s)}$ in $g_s$.

Now if $T'(s)$ is very small, we have $||g||_2 \approx ||g_m||_2$ and keeping $C$ constant is justified. Consider now the case that $T'(s)$ is not small. Since for aligned DPVI we only need to clip and perturb the gradient $g_m$, and because now $||g_m||_2 \leq ||g||_2$, this will lead to less aggressive clipping than in the vanilla case if $C$ is kept constant. It is well known that clipping introduces bias to the solution and we aim to compare cases in which the solution is not changed. Therefore we must adopt different clipping thresholds for the vanilla and the aligned DPVI variants, so that the clipping multiplier $\alpha$ (and hence the gradient after clipping) stays the same. Given some clipping threshold $C'$ for the aligned variant, we derive for the clipping threshold $C$ of the vanilla variant:
\begin{align}
    C &= C' \frac{||g||}{||g_m||} \geq C' \sqrt{1 + T'(s)^2\draw^2}.
\end{align}

Inserting into the variances computed above, we obtain
\begin{align}
    \Var{\draw,\dpdraw}{ \tilde{g}_s } &= T'(\qs)^2 \Var{\draw}{ \draw \alpha \gm \elbo(q) } + \sigma_{DP}^2 C^2 \\
    &\geq T'(\qs)^2  \Var{\draw}{\draw \alpha \gm \elbo(q) } + \sigma_{DP}^2 C'^2 (1 + T'(s)^2) \\
    &= T'(\qs)^2  \left( \Var{\draw}{ \draw \alpha \gm \elbo(q) } + \sigma_{DP}^2 C'^2 \right) + \sigma_{DP}^2 C'^2\text{, and} \\
    \Var{\draw,\dpdraw}{ \tilde{g}_s^{aligned} } &= T'(\qs)^2 \Var{\draw}{ \draw  \alpha \gm \elbo(q) } + T'(\qs)^2 \sigma_{DP}^2 C'^2 \\
    &= T'(\qs)^2 \left(\Var{\draw}{ \draw \alpha \gm \elbo(q) } + \sigma_{DP}^2 C'^2\right).
\end{align}

Therefore $\Var{\draw,\dpdraw}{ \tilde{g}_s } \geq \Var{\draw,\dpdraw}{ \tilde{g}_s^{aligned} }$ holds even for large $T'(s)$ if we adapt $C$ to keep the clipping-induced bias constant.

\section{Variance of DP from OU process on convergence}
\label{app:DPVI_OU}
Directly adapting \citet{MandtHB17}, Sec.~3.2 but ignoring noise from gradient sampling (i.e., assuming full gradients).

\begin{align}
    \xi(t+1) &= \xi(t) - \alpha (\nabla_{\xi} \mathcal{L}(\xi(t)) + (B + \sigma_{DP}I)\eta) \text{ , } \eta \sim \mathcal{N}(0, I) \\
    \Delta \xi(t) &:= \xi(t+1) - \xi(t) = -\alpha \nabla_{\xi} \mathcal{L}(\xi(t)) - \alpha (B+\sigma_{DP}I)\eta \label{eq:OU_finite_diff}
\end{align}

We assume that we can approximate \ref{eq:OU_finite_diff} by stochastic differential equation:
\begin{equation}
    d\xi(t) = -\alpha \nabla_{\xi}\mathcal{L}(\xi(t)) - \alpha (B + \sigma_{DP}I) dW(t)
\end{equation}

Assume that around the optimum $\xi^*$ $\mathcal{L}$ is locally well approximated by a quadratic function
\begin{equation}
    \mathcal{L}(\xi) \approx \frac{1}{2} (\xi-\xi^*)^T A (\xi-\xi^*)
\end{equation}

($A = \frac{\partial^2}{(\partial \xi)^2} \mathcal{L}(\xi^*)=$).

Then we get an Ornstein-Uhlenbeck process
\begin{equation}
    d\xi(t) = -\alpha A (\xi-\xi^*) dt + \alpha (B+\sigma_{DP}) dW(t)
\end{equation}
with stationary distribution
\begin{equation}
    q(\xi) \propto \exp\left\{ -\frac{1}{2}(\xi-\xi^*)^T\Sigma^{-1}(\xi-\xi^*) \right\}
\end{equation}
where $\Sigma$ satisfies
\begin{equation}
    \Sigma A + A \Sigma = \alpha(B+\sigma_{DP}^2 I).
\end{equation}
As we do not have access to the $A$ in general (and we cannot estimate it either since it depends on
the sensitive data), we can still make an important discovery from this Lyapunov equation. Since the $A$
is assumed to be fixed around the optima, we see that the noise covariance of our OU process scales 
linearly with respect to the $\sigma_{DP}^2$.

The above analysis holds for constant learning rate SGD. However the constant learning rate SGD converges poorly, and instead we have used the Adam optimization method \citep{KingmaB14} in our experiments. While Adam does adapt the learning rate, recently \citet{MohapatraSHKT21} showed that the learning rate of Adam will converge to a static value, thus leading to same analysis of OU as above.

\section{Hyperparameters}
\label{app:hyperparameters}

We use Adam \citep{KingmaB14} as the optimiser for all the experiments with starting learning rate of $10^{-3}$. In all of our experiments, the $\delta$ privacy parameter was set to $1/N$ where $N$ denotes the size of the training data.

\paragraph{For the UKB experiment}
In the experiments, we used various different training lengths (depicted e.g. in Figure \ref{fig:num_epochs_test_ukb}). For all of our runs, we set the subsampling rate as $0.01$. The clipping threshold $C$ was set to $C=2.0$ for the aligned and vanilla, $4.0$ for preconditioned variant and to $0.1$ for the natural gradient based variants.

\paragraph{For the Adult experiment}
The training was run for $\numprint{4000}$ epochs with subsampling ratio of $0.01$, corresponding to total of $\numprint{400000}$ gradient steps.

We chose the clipping thresholds for the gradient perturbation algorithm as the $97.5\%$ upper quantile of the training data gradient norms at the non-private optima. This was done to avoid clipping-induced bias, thus making the models comparable to the non-private baseline. This lead to clipping thresholds $C$ presented in Table \ref{tab:adult_clip}.

\begin{table}[h]
    \caption{Clipping thresholds for the Adult data logistic regression model \label{tab:adult_clip}}
    \centering
    \begin{tabular}{ll}
    \hline
    Variant & C \\ \hline
    Aligned & 3.0 \\ \hline
    Aligned Natural Grad. & 0.1 \\ \hline
    Natural Grad. & 0.1 \\ \hline
    Vanilla & 3.0 \\ \hline
    Preconditioned & 4.0 \\ \hline
    \end{tabular}
\end{table}

\section{Model priors}
\subsection{For the UKB experiment}
\label{app:ukb_model_priors}

Recall the probabilistic model used in the UKB experiment:
\begin{align}
    &p(\bX \mid \params_{\bX}, \bpi) = 
            \sum_{k=1}^{K} \pi_k \prod_{j=1}^d \text{Categorical}(\bX_j \mid \params_{\bX}^{(k)}) \\
    &p(\by \mid \bX, \params_{\by}) = \text{Poisson}(\by \mid \exp(\bX \params_{\by})).
\end{align}

The categorical probabilities $\params_{\bX_j}^{(k)}$ for each of the categorical features $\bX_j$, were given a uniform Dirichlet($\mathbf{1}$) prior. Similarly the mixture weights $\boldsymbol\pi$ we assigned a uniform Dirichlet prior. The regression coefficients $\btheta_{\by}$ were given a std. normal $N(0, I)$ prior.

\subsection{For the Adult experiment}
We use the following model and prior
\begin{align}
    \by &\sim \sigma(\bX \bw), \\
    \bw &\sim \mathcal{N}(\boldsymbol{0}, \boldsymbol{I}),
\end{align}
where $\sigma(\cdot)$ denotes the logistic regression function, $\sigma(x) = \nicefrac{1}{1 + \exp{-x}}$.

\section{More results for robustness}
\label{app:more_robustness_results}

\begin{figure}[!tb]
    \centering
    \includegraphics[width=.48\textwidth]{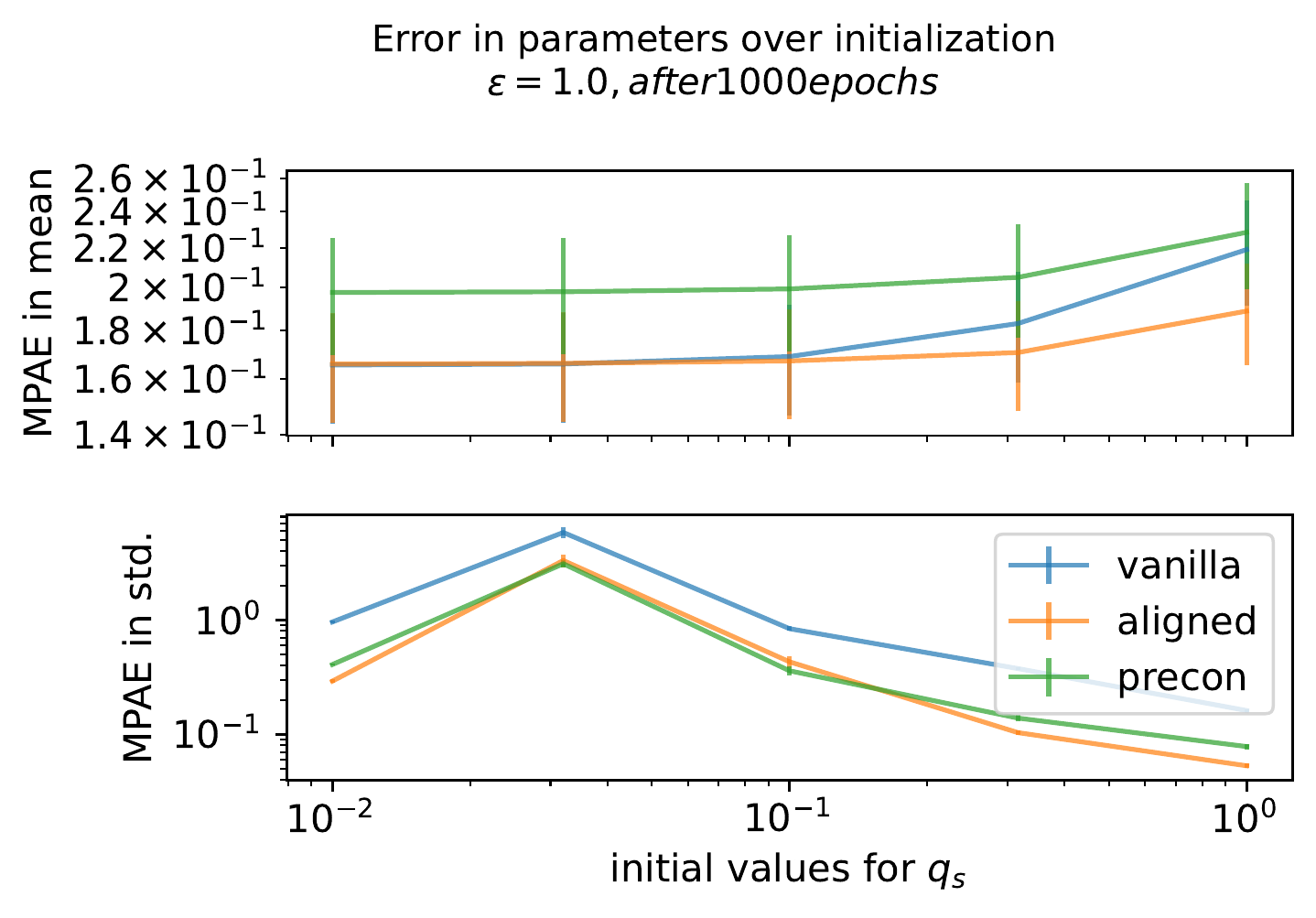}
    \caption{The aligned variant has consistently low error across different initial values of $q_\sigma$. Figure shows the error for $\epsilon=1$ for $1000$ epochs of training. \label{fig:init_test_ukb}}
\end{figure}

\begin{figure}[!tb]
    \centering
    \begin{subfigure}[b]{0.48\columnwidth}
        \includegraphics[width=\columnwidth]{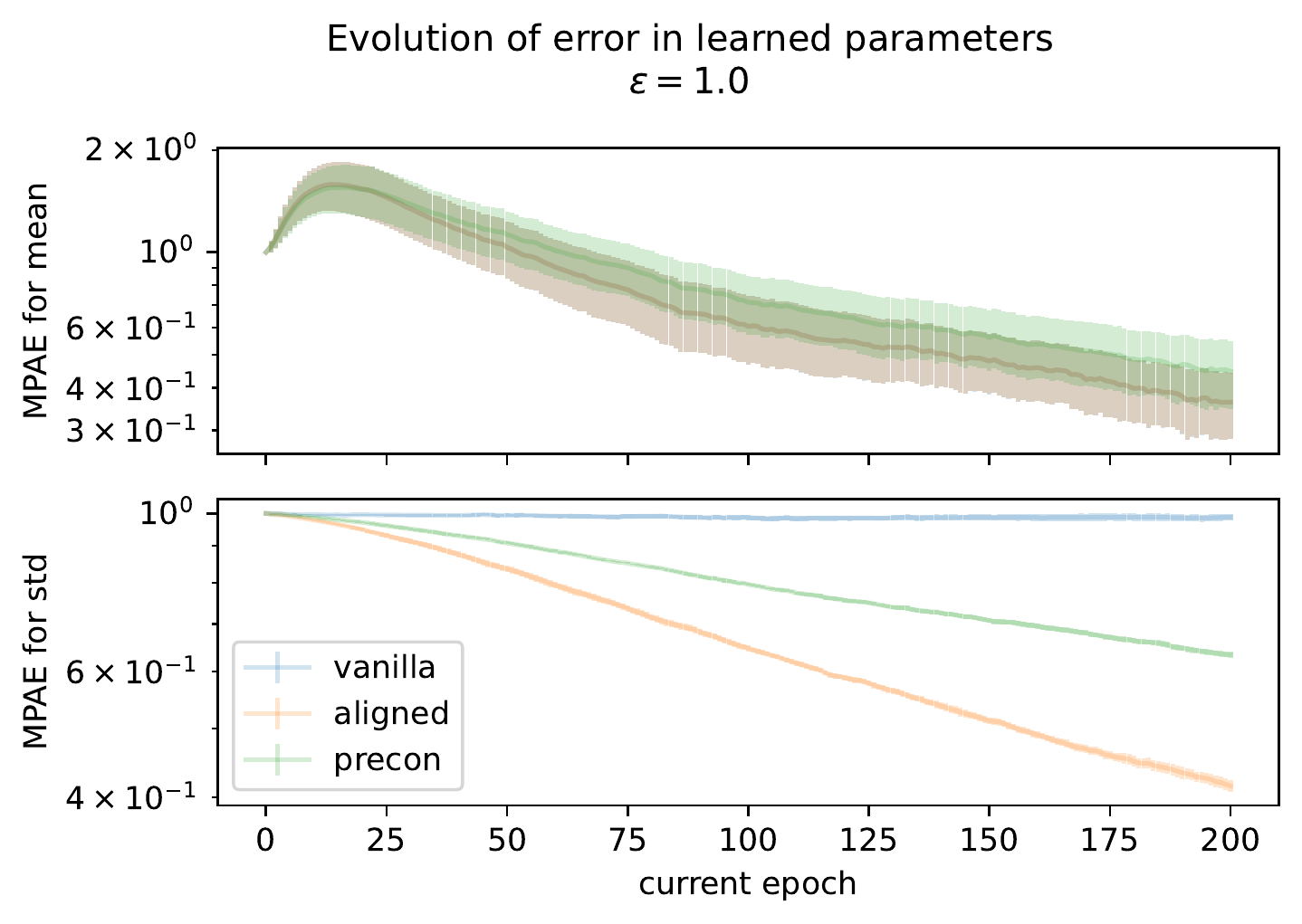}
        \caption{\label{fig:app_init_test_ukb_trace_0.01}}
    \end{subfigure}
    \begin{subfigure}[b]{0.48\columnwidth}
        \includegraphics[width=\columnwidth]{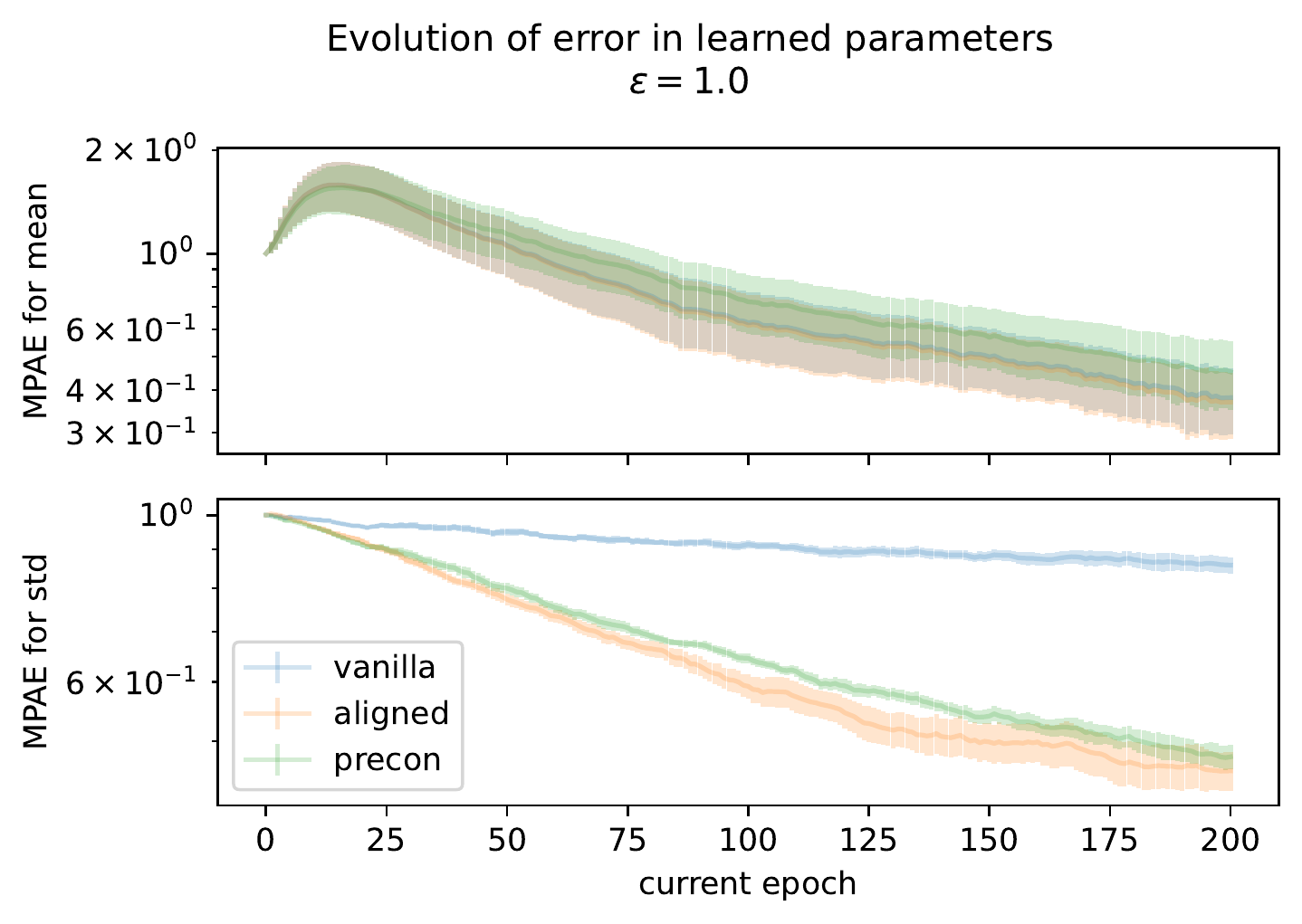}
        \caption{\label{fig:app_init_test_ukb_trace_1.0}}
    \end{subfigure}
    \caption{Aligned DPVI consistently converges faster than the other methods for different initialisation of $\qs$. \textbf{On left}, the $\qs$ is initialised such that $\qsig=0.01$ and \textbf{on right} such that $\qsig=0.1$. \label{fig:app_init_test_ukb_traces}}
\end{figure}

Figure~\ref{fig:init_test_ukb} shows how the MPAE for the different DPVI variants behave for the different initial values for $\qs$ for both, variational mean (upper panel) and standard deviation (lower panel) after $\numprint{1000}$ epochs.

Figure~\ref{fig:app_init_test_ukb_traces} shows the parameters traces for $\qs$ initialised such that $\qsig=0.01$ (left) and $\qsig=0.1$ (right) with the same split in upper and lower panels, similar to Figure~\ref{fig:init_test_ukb_trace_1.0} for $\qsig=1.0$ in the main body of the paper.

We observe that with decreasing initial values for $\qs$ (/$\qsig$) it becomes increasingly difficult for vanilla DPVI to learn variational standard deviation but learning of means is slightly improved. Preconditioned DPVI performs better overall in terms of standard deviation but learns means worse. Aligned DPVI consistently outperforms both competing variants.

\subsection{Natural gradients and the aligned natural gradients do not converge}

\begin{figure}[tb]
    \centering
    \includegraphics[width=.48\textwidth]{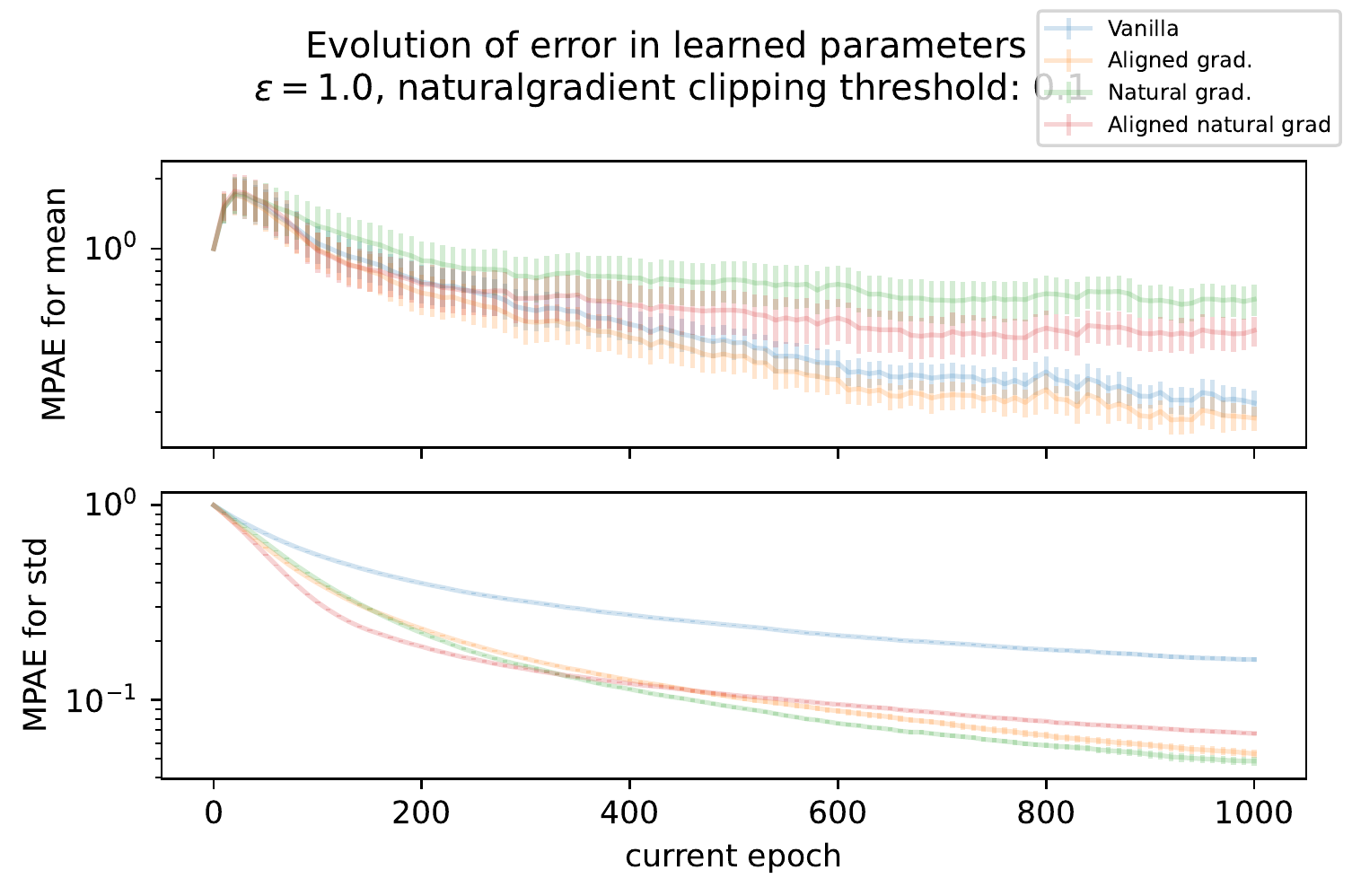}
    \caption{The natural gradient based variants struggle to converge in the UKB experiment. Figure shows the evolution of MPAE for both of the variational parameters over 1000 epochs. Lines show the mean MPAE over 10 independent repeats as well as std. of mean as error. In this experiment, the $\qsig$ was initialised to 1 and clipping threshold for both the natural gradient variants was set to 0.1. \label{fig:app_ng_ukb}}
\end{figure}

Besides the vanilla and aligned variant, we also fitted the UKB model using the natural gradient and aligned natural gradient variants. From Figure \ref{fig:app_ng_ukb} we can see that both the natural gradient and aligned natural gradient variants converge more slowly than the aligned variant. We can again see the trade-off natural gradient makes; the means are learned worse than standard deviations, which is what we expect based on the analysis of Section \ref{sec:dpvi_variants}. Somewhat surprisingly, the aligned natural grad.~variant performs worse than the natural grad.~in this experiment. This might be due to poor choice of hyperparameter, for example the learning rate for the Adam optimiser used in the experiments was set to $10^{-3}$ for all the variants, while we know that the natural grad.~variants tend to have smaller gradients than the others - although Adam should in theory be able to adapt to that.

\begin{figure}[tb]
    \centering
    \begin{subfigure}[t]{.48\textwidth}
        \includegraphics[width=\textwidth]{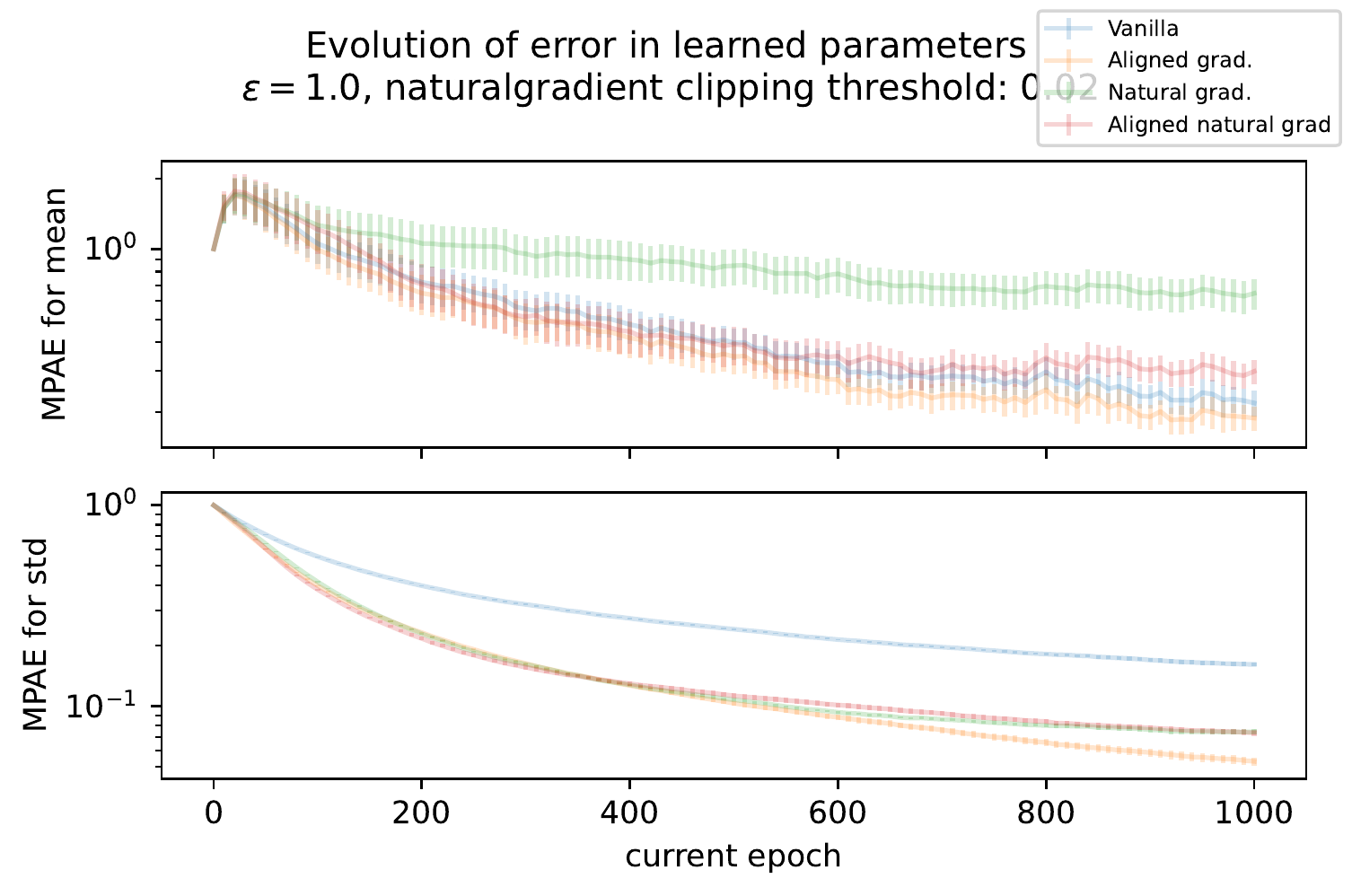}
        \caption{The aligned natural gradient variant performs slightly better when the clipping threshold is set to $C=0.02$. The natural gradient variant appears to diverge from the true variational std. which might be caused by clipping-induced bias, while still struggling to learn the correct variational mean. \label{fig:less_clipped_ng_002}}
    \end{subfigure}
    \hfill
    \begin{subfigure}[t]{.48\textwidth}
        \includegraphics[width=\textwidth]{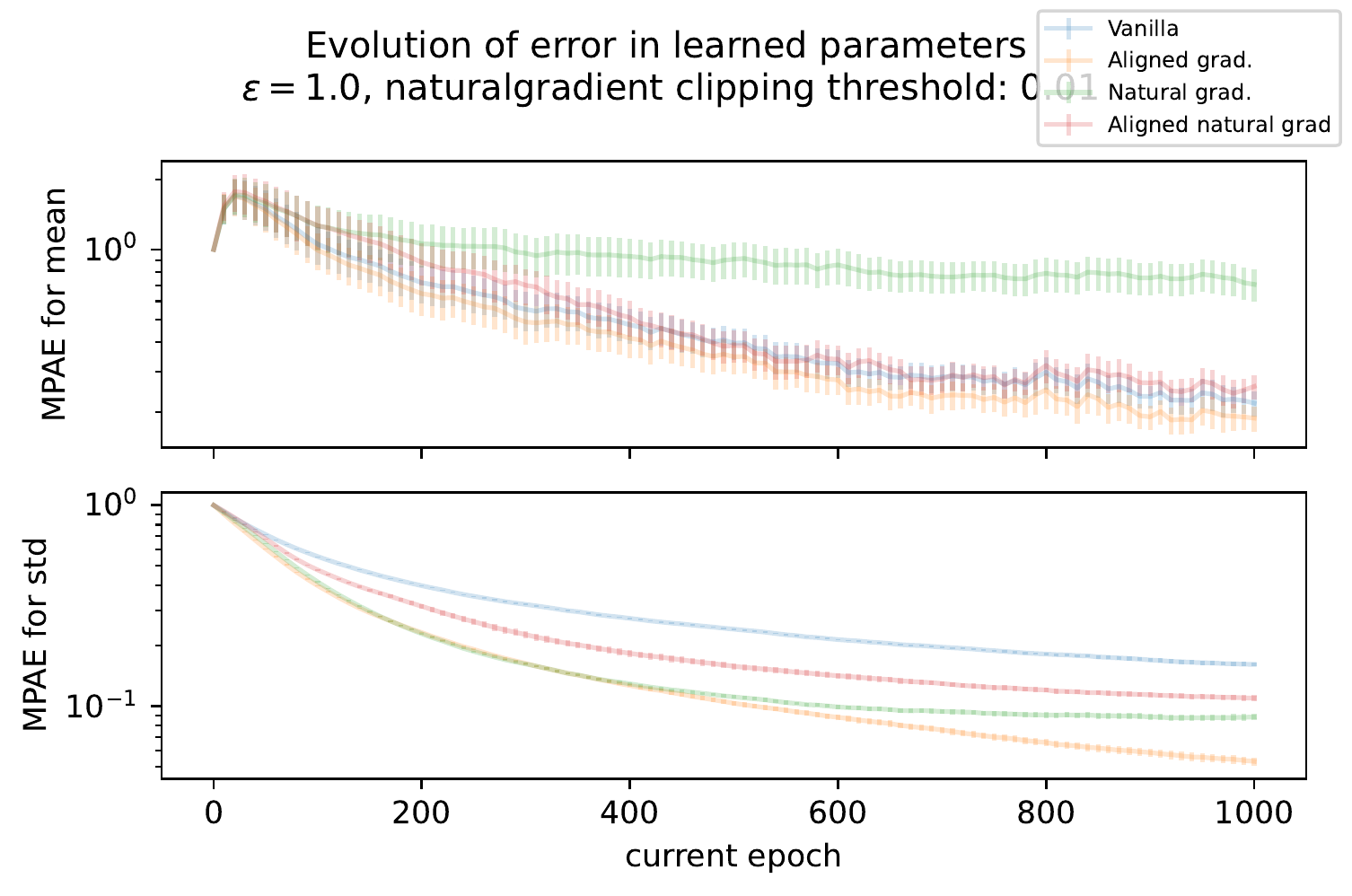}
        \caption{Both natural gradient variants start to diverge if clipping is set too low $C=0.01$. \label{fig:less_clipped_ng_001}}
    \end{subfigure}
    \caption{Tests with smaller clipping threshold for the natural gradient variants. The clipping threshold for the vanilla and aligned is still set to $2.0$. \label{fig:less_clipped_ng}}
\end{figure}

We also tried decreasing the clipping threshold for the natural gradient variants to see if the slower convergence is due to a too high level of DP noise. Figure \ref{fig:less_clipped_ng_002} shows that reducing the clipping threshold helps the aligned natural gradient variant to converge in the variational std's. However, the method still converges slower in the variational means. We suspect that this is due to Adam not being able to adjust the learning rate fast enough when learning the means. This is because the gradients of means become scaled with the variational std's in the natural gradient approach and become small when the posterior distribution begins to converge. Figure \ref{fig:less_clipped_ng_001} shows that setting clipping threshold $C=0.01$ natural gradient variants is too small, and the aligned natural gradient starts to suffer from clipping-induced bias.

\section{Further details on downstream analysis for UKB data}
\label{app:downstream_details}
In the UKB experiment, we use the learned variational posterior to sample a synthetic data set from the posterior predictive distribution (PPD) as suggested by \citet{JalkoLHTHK21}. We test the method by comparing the synthetic data in downstream analysis to the original data. As the downstream task, we fitted a Poisson regression model that aims to predict whether individual catches SARS-CoV-2 based on the predictors in the data. Note that this downstream perfectly overlaps with our generative model.

In order to properly reflect the uncertainty rising from the data generating process to the final results computed from the synthetic data, we will employ so called \emph{Rubin's rules} \citep{Rubin04}. In this procedure, we first sample multiple synthetic data sets from the PPD and compute the downstream analysis on each of the sampled synthetic data. Next, the results are aggregated according to a set of rules and we recover finally a more robust estimator for our downstream analysis. Further discussion about the Rubin's rules can be found for example in \citep{Reiter07}.

In our experiments, we sampled $100$ data sets from the PPD learned using the aligned variant, and applied the Rubin's rules to compute a mean and std. estimate for the Poisson regression coefficients. Finally, the obtained means were compared to the Poisson regression coefficients learned using the original data.

\section{Experimental setup for full-rank Gaussian approximation}
\label{app:full_rank_experiment}

In this experiment we create simulated data where we control the amount of correlations between data dimensions as the ratio $\rho$ of non-zero off-diagonal entries in the correlation matrix. To generate data with $d$ dimensions and correlation density $\rho$, we
\begin{enumerate}
    \item generate a correlation matrix $\boldsymbol{C}$ using Algorithm~\ref{alg:sample_corr_matrix} with inputs $d, \rho, \alpha=8, \beta=10$,
    \item sample a diagonal matrix $\boldsymbol{D}$ of marginal variances, where $\{D\}_{ii} \sim \exp{\mathcal{N}(0, 0.2^2)}$,
    \item obtain the covariance matrix $\boldsymbol{\Sigma} = \boldsymbol{D} \boldsymbol{C} \boldsymbol{D}$
    \item sample $N = \numprint{10000}$ data points $\bx_n \sim \mathcal{N}(\boldsymbol{0}, \boldsymbol{\Sigma})$
    \item sample random regression weight vector $\bw \sim \mathcal{N}(\boldsymbol{0}, \boldsymbol{I})$
    \item sample $\by \sim \mathcal{N}(\bX \bw, \sigma_y^2)$, with $\sigma_y = 1$.
\end{enumerate}

We perform the above for all combinations of $d = 100, 200$ and $\rho = 0.2, 0.8$. We then use vanilla DPVI and DPVI with aligned gradients to learn the full-rank Gaussian posterior approximation to the Bayesian linear regression model with priors
\begin{align}
    \by &\sim \mathcal{N}(\bX \bw, \sigma_y^2), \\
    \bw &\sim \mathcal{N}(\boldsymbol{0}, \boldsymbol{I}) \\
    \sigma_y &\sim \text{Gamma}(0.1, 0.1).
\end{align}

We run the inference for $\numprint{1000}$ epochs, gradient clipping threshold $0.2$ and subsampling ratio $0.01$.

For the same $d, \rho$ we then generate another $\numprint{10000}$ data points and compute the log-likelihood using the obtained posterior approximation. We repeat the inference and evaluation $\numprint{50}$ times for each method and combination of $d$ and $\rho$, keeping the generated training and testing set fixed.

\begin{algorithm}[h]
\begin{algorithmic}
\Require $d, \rho \in [0,1], \alpha > 0, \beta > 0$
\Ensure correlation matrix $\boldsymbol{C}$
\State $K \gets \rho \frac{d(d-1)}{2}$ \Comment{number of non-zero off-diagonal entries}
\State $U \gets \emptyset$
\State $k \gets 0$
\State $\boldsymbol{C} \gets \boldsymbol{I}_d$
\For {$k = 1, \ldots, K$}
\State sample $(i,j) \in T \setminus U$ at random
\State sample $c \sim \text{Beta}(\alpha, \beta)$ \Comment{sample correlation strength, controlled by $\alpha$ and $\beta$}
\State sample $f \in \{-1, 1\}$ at random \Comment{sample sign of correlation}
\State $C_{ij} \gets f c$
\State $C_{ji} \gets C_{ij}$
\State $U \gets U \cup \{(i,j),(j,i)\}$
\EndFor
\end{algorithmic}
\caption{Routine to generate a $d$-dimensional correlation matrix with given density $\rho$ and strength of correlations controlled by $\alpha, \beta$.}
\label{alg:sample_corr_matrix}
\end{algorithm}

\section{Runtimes}
\label{app:runtime}

\subsection{UKB experiments}
In this experiment, we ran all the variants separately for 10 seeds and 4 levels of privacy.
Additionally, we experimented with different runtimes and initialisations. For a training of \numprint{1000} epochs, a single repeat takes between 20 to 40 minutes. The runtime scales linearly with the number of epochs.

Further, we computed the downstream task for the aligned variants, which includes generating the 100 synthetic data sets and fitting the downstream Poisson regression model on those 100 synthetic data. This procedure takes between 30 to 60 minutes to complete.

A rough estimate of the runtimes for the UKB experiment is given in Table \ref{tab:ukb_runs}. A single CPU core with 8gb of memory was used for all the runs.

\begin{table*}[t]
    \centering
    \caption{Estimated runtimes for UKB experiment\label{tab:ukb_runs}}
    \begin{tabular}{lllll}
    \hline
    \#epochs & single repeat runtime & repeats & \# epsilon values & \# initial values \\ \hline
    200 & 4-8min & 10 & 4 & 5 \\ \hline
    400 & 8-16min & 10 & 4 & 1 \\ \hline
    600 & 12-18min & 10 & 4 & 1 \\ \hline
    800 & 16-32min & 10 & 4 & 1 \\ \hline
    1000 & 20-40min & 10 & 4 & 1 \\ \hline
    2000 & 40-80min & 10 & 4 & 1 \\ \hline
    4000 & 80-160min & 10 & 4 & 1 \\ \hline
    8000 & 160-320min & 10 & 4 & 1 \\ \hline
    \end{tabular}
\end{table*}

\subsection{Adult experiments}
A single training repeat of learning the logistic regression model for all the different variants of DPVI, took between 10 and 30 minutes to finish on a single CPU core with 8gb of memory assigned. In total, the Adult experiment was repeated 50 times for four different levels of privacy. Therefore the total runtime of all the experiments is \textbf{between \numprint{2000} and \numprint{6000} minutes}.

The variance in running times is likely due to differences in computation nodes in the clustered assigned by a automatic run scheduler.

\subsection{Full-rank experiments}
A single run for this experiment consisted of the inference using both vanilla and aligned DPVI with full-rank approximations. All runs were executed on a computing cluster utilising Nvidia K80, A100, P100, V100 GPU hardware, to which the runs were allocated automatically to balance overall load. As a result, runtimes varied slightly: Runs for and $\numprint{100}$ dimensional data took 6-8 minutes to finish, runs for $\numprint{200}$ dimensions took 8-10 minutes. With a total of 4 data set configurations and 50 repeats for each, the total runtime is $\numprint{1400}$ to $\numprint{1800}$ minutes.

\section{Gradient distributions for different variants}

\begin{figure}[tb]
    \centering
    \begin{subfigure}{.32\textwidth}
        \includegraphics[width=\textwidth]{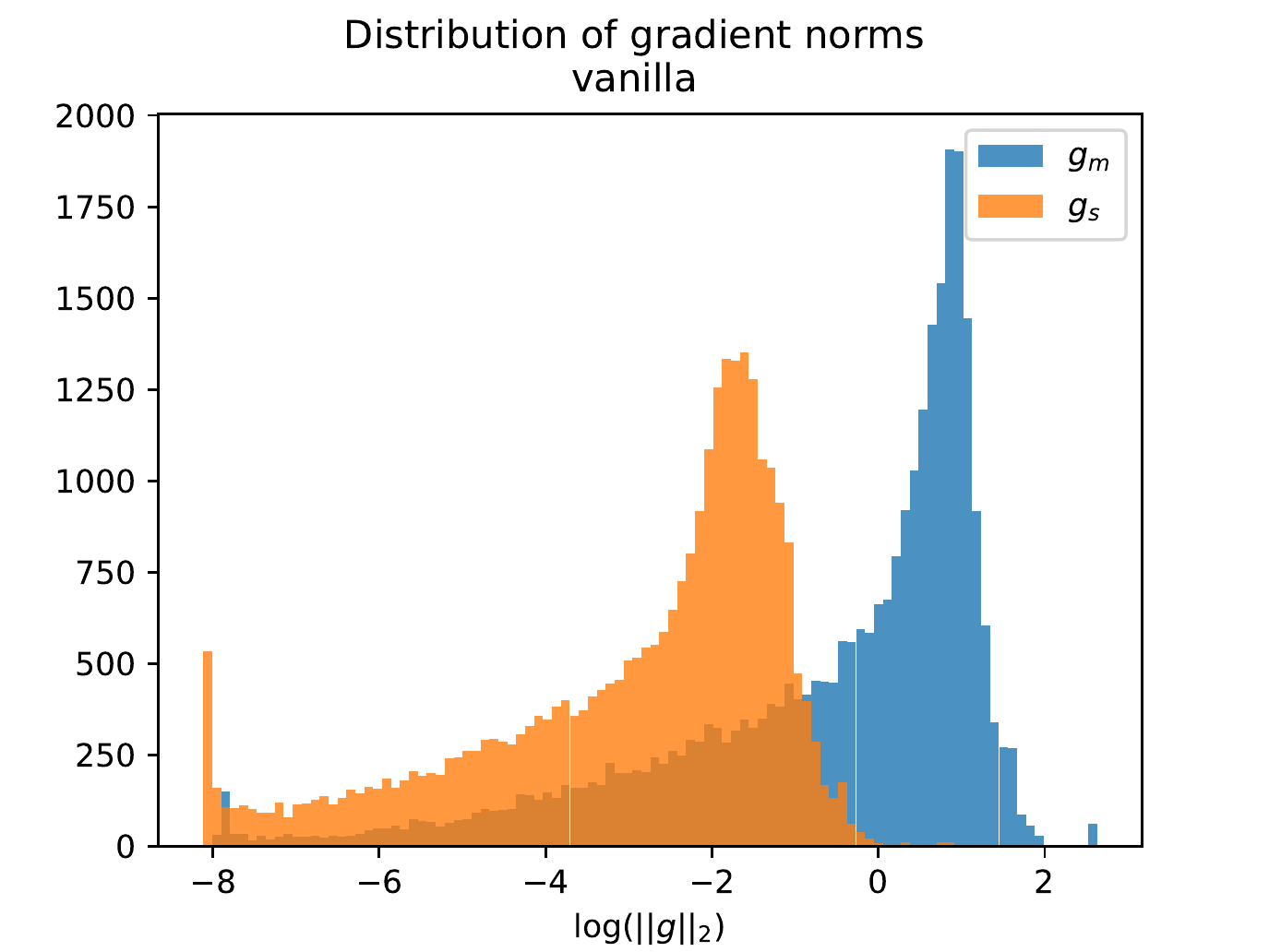}
        \caption{Vanilla gradients} \label{fig:grad_dists_vanilla}
    \end{subfigure}
    \begin{subfigure}{.32\textwidth}
        \includegraphics[width=\textwidth]{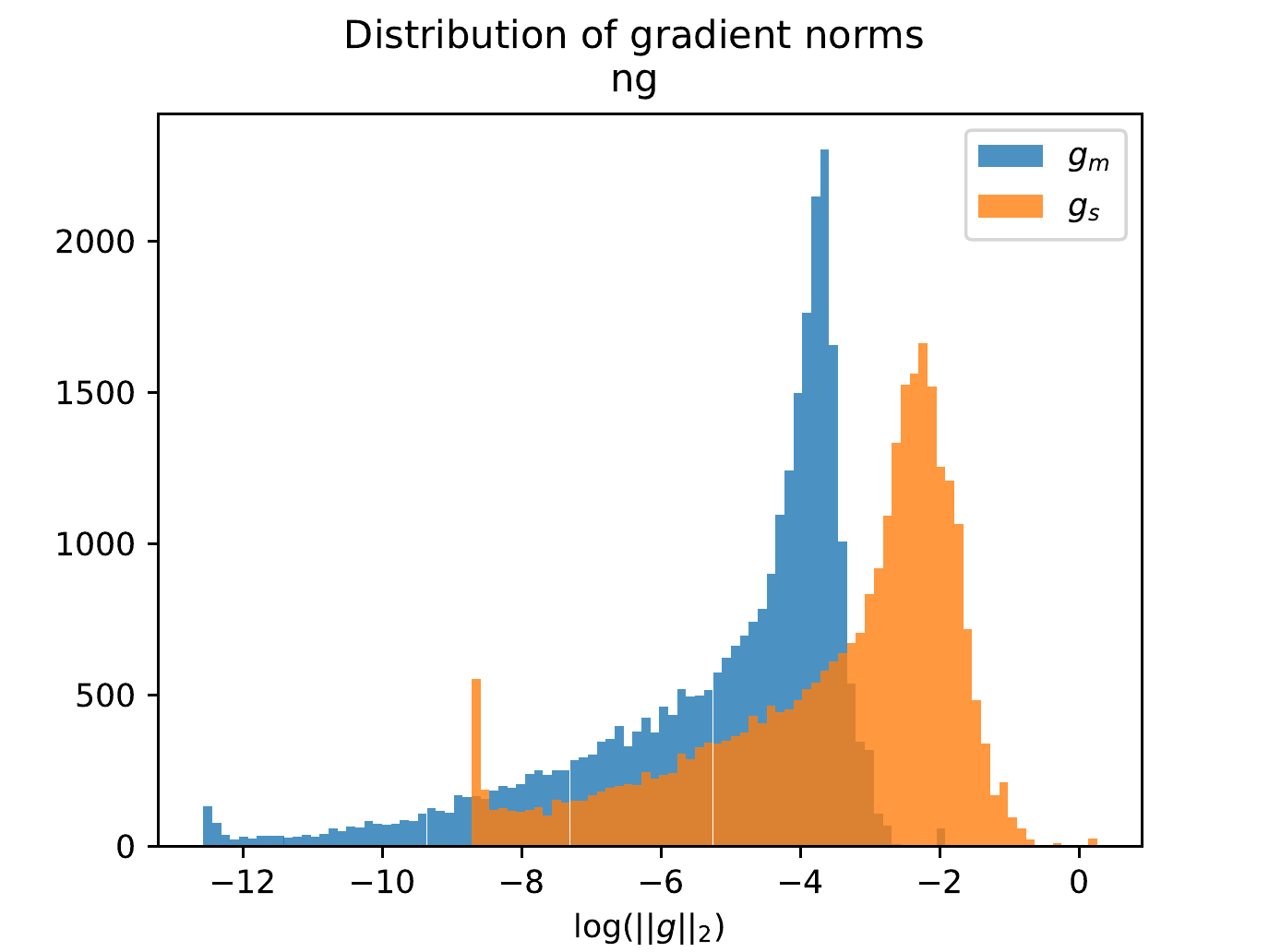}
        \caption{Natural gradients} \label{fig:grad_dists_ng}
    \end{subfigure}
    \begin{subfigure}{.32\textwidth}
        \includegraphics[width=\textwidth]{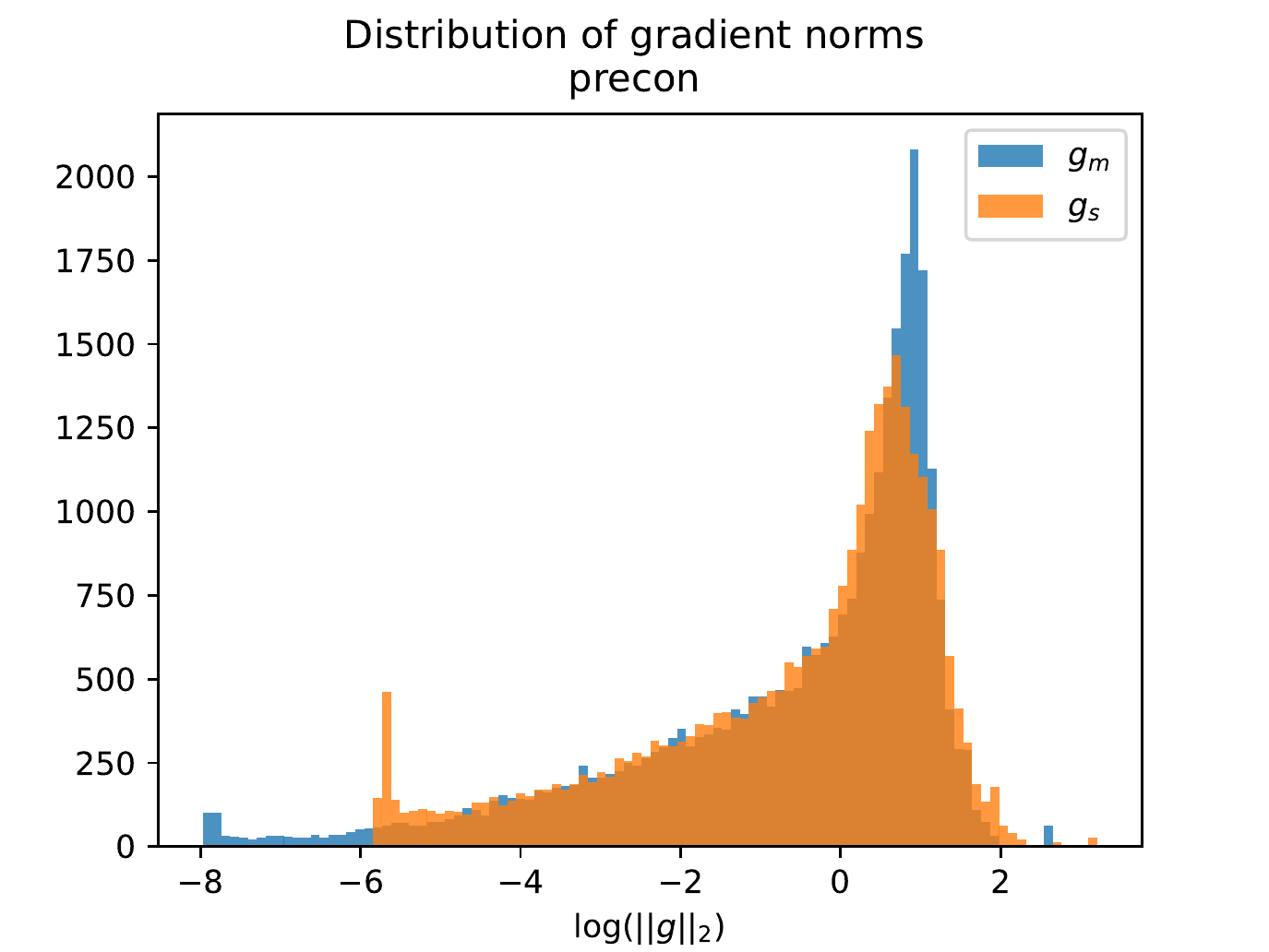}
        \caption{Preconditioned gradients} \label{fig:grad_dists_precon}
    \end{subfigure}
    \caption{\label{fig:grad_dists}}
\end{figure}

Figure~\ref{fig:grad_dists} shows the distributions of gradient norms for variational means and scales for different variants of DPVI discussed in Section~\ref{sec:dpvi_variants} when $\qs$ is set to $0.1$. Figure~\ref{fig:grad_dists_vanilla} clearly shows the different magnitudes for variational standard deviation in vanilla DPVI. Figure~\ref{fig:grad_dists_ng} demonstrates that natural gradients simply reverse the problem. Figure~\ref{fig:grad_dists_precon} shows that the scaling approach achieves matching magnitudes quite well. However, it comes at the cost of increasing the norm of the full (combined) gradient and therefore increased sensitivity.

\printbibliography[heading=subbibliography, filter=appendixOnlyFilter]

\end{document}